\documentclass[10pt]{article}

\PassOptionsToPackage{numbers, compress}{natbib}

\usepackage[preprint]{neurips_2026}

\usepackage{amsmath,amssymb,amsfonts}
\usepackage{algorithm}
\usepackage{algorithmic}
\usepackage{booktabs}
\usepackage{graphicx}
\usepackage{hyperref}
\usepackage{microtype}
\usepackage{pifont}
\usepackage[most]{tcolorbox}
\usepackage{xcolor}
\usepackage{pgfplots}
\pgfplotsset{compat=1.16}
\usepackage{colortbl}

\newcommand{\cmark}{\ding{51}}%
\newcommand{\xmark}{\ding{55}}%

\title{When Load-Balancing Goes Too Far: Expert Pruning\\in Over-Dispersed Mixture-of-Experts Models}

\author{
  \bf Berkcan Kapusuzoglu\thanks{Correspondence to \texttt{berkcan.kapusuzoglu@capitalone.com}.} \quad Connor Pryor \quad Sangwoo Cho\\
  \bf Supriyo Chakraborty \quad Shi-Xiong Zhang \quad Sambit Sahu \quad Milind Naphade\\
  Capital One, AI Foundations, McLean, VA 22102, USA
}

\date{}

\begin{document}
\maketitle

\begin{abstract}
Expert pruning reduces the memory and serving cost of Mixture-of-Experts (MoE) models by removing low-importance experts identified by the router. This approach assumes that router probabilities provide a reliable importance signal. We observe that this assumption breaks down under \emph{over-dispersed routing}, a regime associated with aggressive load-balancing during training, in which tokens are distributed nearly uniformly across experts and importance signals collapse. In this regime, perplexity does not predict downstream task accuracy. On gpt-oss-20B, the lowest-perplexity pruning configuration yields the worst mathematical reasoning performance, while the highest-perplexity configuration preserves it. This behavior does not occur under standard routing (e.g., Mixtral-8x7B-Instruct), where perplexity and accuracy degrade together as expected. We further show that under over-dispersed routing, pruning exposes a capability trade-off: no single scoring metric dominates across tasks. Activation-aware scoring preserves mathematical reasoning but severely degrades knowledge-intensive science tasks (an 18\% gap on GPQA), whereas frequency-based scoring exhibits the reverse. To navigate this trade-off, we propose Minimax Expert Score Allocation (MESA), a domain-aware pruning method that iteratively boosts importance scores for experts serving whichever domain is currently worst-affected, minimizing the worst-case domain degradation rather than optimizing average accuracy. At 25\% expert pruning, MESA achieves the smallest worst-case degradation across domains, including an 18-point GPQA gap that activation-aware scoring incurs; on a per-benchmark basis, this translates to MESA outperforming activation-aware baselines on 7 of 11 benchmarks, at a correspondingly reduced memory footprint. This trade-off does not arise under standard routing, where all methods degrade proportionally and no domain-specific imbalance emerges. Our results indicate that over-dispersed routing is a qualitatively distinct pruning regime in which standard assumptions fail, and that recognizing this regime is a prerequisite for principled expert pruning of load-balanced MoE models.
\end{abstract}

\section{Introduction}
\label{sec:intro}

Mixture-of-Experts (MoE) architectures~\citep{shazeer2017outrageously, fedus2022switch, zoph2022stmoe} achieve favorable compute--quality tradeoffs by activating only a subset of parameters per token. However, the large total parameter count of MoE models creates substantial memory and serving costs. \emph{Expert pruning}, the structured removal of entire expert modules, is a natural strategy for reducing these costs~\citep{lu2024not,he2024demystifying,liu2026evoesapnonuniformexpertpruning}, and recent work has shown it outperforms expert merging~\citep{li2024mcsmoe, chen2025retrainingfreemergingsparsemoe} on generative tasks by preserving router-controlled specialization~\citep{lasby2026reapexpertspruningprevails}.

Prior work on expert pruning has focused on models with well-balanced routing distributions, where expert importance varies meaningfully and low-importance experts can be safely removed~\citep{lu2024not, huang2025shapleymoe}. We identify a distinct regime that we term \emph{over-dispersed routing}: models trained with unusually high auxiliary load-balancing loss coefficients (e.g., $\lambda_{\text{aux}}=0.9$). In over-dispersed models, the router distributes tokens nearly uniformly across all experts, creating two challenges for pruning:

\begin{enumerate}
\item \textbf{Perplexity (PPL)--accuracy disconnect.} Standard perplexity-based evaluation is unreliable for pruned over-dispersed models. Across five pruning configurations, we observe that the expected negative correlation between PPL and accuracy (lower PPL $=$ better) does not hold: Spearman $\rho$ is positive on GSM8K ($\rho=0.70$, $p > 0.1$) and near zero on GPQA ($\rho=-0.10$). The lowest-PPL configuration (random pruning, PPL$=$40.95) produces the worst task accuracy, while the highest-PPL method (REAP, PPL$=$145.04) retains strong performance. This is the opposite of standard pruning behavior, where PPL and accuracy are observed to degrade together (Section~\ref{sec:ppl}).

\item \textbf{Domain unfairness.} While macro-level routing appears near-uniform, experts retain subtle domain-specific characteristics that are critical for performance. Global importance scores average across all tokens, cancelling out these weak domain signals (as we visualize in Appendix~\ref{app:routing-viz}). Consequently, pruning based on global metrics degrades some domains disproportionately.
\end{enumerate}

Over-dispersed routing is not a corner case: production MoE systems increasingly use high auxiliary loss coefficients to ensure load balance across hardware~\citep{zoph2022stmoe, fedus2022switch}, and recent open-weight releases (gpt-oss-20b with $\lambda_{\text{aux}}=0.9$~\citep{openai2025gptoss120bgptoss20bmodel}) ship models in this regime. As these models are deployed at scale, practitioners need pruning methods that account for the routing distribution rather than assuming standard expert specialization.

To navigate this trade-off landscape, we propose Minimax Expert Score Allocation (MESA), a domain-fair scoring method. The algorithm iteratively boosts importance scores for experts serving whichever domain is currently worst-affected by the pruning plan, reducing the loss of the worst domain. We evaluate on gpt-oss-20b, and validate cross-architecture on Mixtral-8x7B-Instruct~\citep{jiang2024mixtral}.

Our contributions:
\begin{itemize}
\item We identify and characterize the over-dispersed routing regime ($\lambda_{\text{aux}} \gg 0.01$), in which we observe that the expected negative PPL--accuracy relationship breaks down (Spearman $\rho$ positive or near zero across tasks) and standard importance signals become unreliable. Cross-architecture experiments confirm this association is regime-specific: the behavior does not occur on Mixtral-8x7B-Instruct under standard routing.
\item We demonstrate a capability trade-off not examined in prior work: under over-dispersed routing, no single pruning metric dominates across tasks. Activation-aware methods (REAP) preserve competition math (AIME~2025 73\%) but lose science tasks (GPQA 47.72\% vs.\ 65.48\%, an 18\% gap), while frequency-based methods exhibit the reverse.
\item We propose MESA, a minimax domain-fair scoring algorithm that validates this characterization by explicitly minimizing the worst-affected domain's degradation rather than average-case accuracy: it achieves the smallest worst-case degradation across domains at $r=0.25$, which on a per-benchmark basis corresponds to outperforming REAP on 7 of 11 benchmarks (5 statistically significant).
\end{itemize}

\section{Related Work}
\label{sec:related}

\textbf{Expert importance scoring.}\quad REAP~\citep{lu2024not} scores experts by the product of routing probability and activation $\ell_2$ norm, capturing both selection frequency and output magnitude. HEAPr extends this with hardware-aware expert placement. DiEP~\citep{he2024demystifying} demonstrates that simple router probability mass is competitive with more complex scores, and identifies layer-level sensitivity as a key factor. The concurrent SHAPE framework~\citep{huang2025shapleymoe} reports 82.44\% average accuracy on gpt-oss-20b at 20\% expert pruning; however, its REAP baseline sets \texttt{expert\_norms=1.0}, effectively reducing REAP to router probability alone and underestimating the true REAP signal. Our REAP implementation computes actual activation $\ell_2$ norms.

\textbf{Expert pruning and allocation.}\quad EvoESAP~\citep{liu2026evoesapnonuniformexpertpruning} uses a genetic algorithm to jointly optimize per-layer pruning counts, achieving state-of-the-art results on Mixtral and DeepSeek-V2. Uniform allocation (same count per layer) and global ranking (score all experts jointly) are common baselines. Recent work reveals a trade-off between multiple-choice (MC) and generative benchmark performance: expert merging methods~\citep{li2024mcsmoe,chen2025retrainingfreemergingsparsemoe} excel on MC tasks (ARC, MMLU, HellaSwag), while pruning better preserves generative capabilities such as mathematical reasoning and code generation~\citep{lasby2026reapexpertspruningprevails}. This MC--generative split motivates evaluation suites that span both task types.

\textbf{Domain fairness in pruning.}\quad Most pruning methods optimize a single global metric (PPL or average task accuracy), which can mask severe degradation on minority domains. Minimax fairness has been applied to group-level training objectives~\citep{sagawa2020distributionally} and to data selection during dense model pruning~\citep{deng2025drpruning}, but not to expert importance scoring in MoE models. MESA adapts the minimax principle to operate on expert scores rather than training losses, requiring no retraining.

\textbf{Over-dispersed routing and load-balancing.}\quad The effect of auxiliary loss magnitude on MoE routing has been studied in the context of training stability~\citep{fedus2022switch,zoph2022stmoe}. \citet{omi2025simbal} show that SimBal rebalancing during fine-tuning can recover expert specialization lost to aggressive auxiliary losses. However, the implications for post-training pruning remain unexplored: when routing is near-uniform, standard importance signals degenerate and pruning decisions become arbitrary. MESA addresses this by incorporating domain-level feedback into the scoring loop, providing a post-training solution that does not require modifying the training pipeline.

\textbf{Domain-fair compression.}\quad \citet{deng2025drpruning} introduce DRPruning, which applies distributionally robust optimization at the \emph{data} level during structured pruning of dense models, upweighting underrepresented domains during fine-tuning. MESA operates at the \emph{expert} level: it adjusts expert importance scores rather than training data weights, requiring no fine-tuning and applying to frozen post-training models.

\section{Method}
\label{sec:method}

We describe MESA, our proposed domain-fair expert pruning method.

\subsection{Over-Dispersed Routing and Its Consequences}
\label{sec:overdispersed}

Consider a standard MoE layer with $E$ experts and top-$k$ routing. Let $\pi_i(x) = \text{softmax}(\mathbf{W}_r \mathbf{x})_i$ denote the routing probability for expert $i$ on input $\mathbf{x}$. The auxiliary load-balancing loss~\citep{shazeer2017outrageously} is:
\begin{equation}
\mathcal{L}_{\text{aux}} = \lambda_{\text{aux}} \cdot E \cdot \sum_{i=1}^{E} f_i \cdot \pi_i
\end{equation}
where $f_i$ is the fraction of tokens routed to expert $i$ and $\pi_i$ is the mean routing probability for expert $i$.

In gpt-oss-20b, $\lambda_{\text{aux}} = 0.9$, orders of magnitude larger than typical values (${\sim}0.001$--$0.01$ in Mixtral~\citep{jiang2024mixtral} and Switch Transformer~\citep{fedus2022switch}). This forces routing probabilities toward uniformity ($\pi_i \approx 1/E$ for all $i$), which we term \emph{over-dispersed routing}.

\textbf{Consequence 1: Pruning can improve local metrics.}\quad When routing is near-uniform with top-$k$ selection, pruning low-contribution experts concentrates routing mass onto the remaining (potentially more specialized) experts. This can \emph{decrease} perplexity while degrading downstream task performance, explaining the PPL--accuracy disconnect.

\textbf{Consequence 2: Global importance scores cancel out micro-signals.}\quad While macro-level routing appears perfectly uniform (e.g., each of the 32 experts in gpt-oss-20b receives ${\sim}3.1\%$ of the global traffic), experts retain subtle domain-specific affinities. For example, the Math domain might route $3.8\%$ of its tokens to Expert~A, while the Code domain routes only $2.4\%$. A global importance score averages these micro-signals together, yielding the baseline $3.1\%$ uniform prior and completely destroying the domain-specific signal (visualized via heatmaps in Appendix~\ref{app:routing-viz}). Consequently, pruning based on global scores accidentally removes critical domain specialists because they appear to be replacable generalists.

The critical distinction between over-dispersed and normal routing is visible not in \emph{aggregate} statistics (which can be misleading) but in \emph{per-token} routing entropy. For each token $\mathbf{x}$, we compute the entropy of the full router softmax distribution: $H(\mathbf{x}) = -\sum_{i=1}^{E} \pi_i(\mathbf{x}) \log_2 \pi_i(\mathbf{x})$. We then average over calibration tokens from all 6 domains.

\textbf{Mixtral-8x7B-Instruct}~\citep{jiang2024mixtral} ($\lambda_{\text{aux}} \approx 0.001$, $E=8$, top-2): The mean per-token entropy is $\bar{H}_{\text{token}} = 2.23$ bits, only 74.3\% of the theoretical maximum $\log_2(8) = 3.0$ bits. Per-token routing is \emph{peaked}: each token's softmax concentrates mass on 2--3 experts, with the remaining experts receiving negligible probability. Entropy varies across depth (early layers ${\sim}80\%$ of max, late layers ${\sim}65\%$) and across tokens within a layer (per-layer std $0.28$--$0.60$ bits). The high token-level variance indicates that \emph{different} tokens specialize to \emph{different} experts; aggregate routing appears balanced because specialization averages out, not because individual routing decisions are uniform.

\begin{figure}[t]
\centering
\includegraphics[width=0.75\columnwidth]{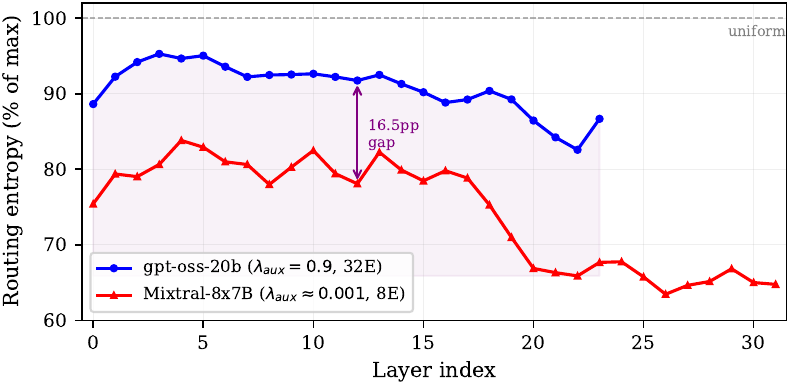}
\caption{Per-layer routing entropy (normalized to \% of maximum) for
gpt-oss-20b and Mixtral-8x7B-Instruct. gpt-oss maintains $>$82\% across all layers
(over-dispersed), while Mixtral drops to 64--68\% in later layers as
specialization emerges. This gap explains why global importance scores are
informative for Mixtral but noisy for gpt-oss.}
\label{fig:entropy-profile}
\end{figure}

\textbf{gpt-oss-20b} ($\lambda_{\text{aux}} = 0.9$, $E=32$, top-4): Mean per-token entropy is $\bar{H}_{\text{token}} = 4.54$ bits (90.8\% of $\log_2(32) = 5.0$ bits), measured over 183K tokens. As shown in Figure~\ref{fig:entropy-profile}, gpt-oss maintains $>$82\% entropy across all layers while Mixtral drops to 64--68\%. Every expert receives approximately $k/E = 4/32 = 12.5\%$ of routing probability \emph{per token}, regardless of token content.

This per-token view explains why standard pruning metrics fail on gpt-oss: in Mixtral, each token has clear ``important'' and ``unimportant'' experts (low $H$ per token $\Rightarrow$ high signal), so global importance scores reliably identify dispensable experts. In gpt-oss, no expert is distinctly unimportant for any individual token (high $H$ per token $\Rightarrow$ low signal), making global scores noisy and domain-conditional measurement essential. Pruning in this regime removes experts and forces the router to redistribute mass to the remaining experts, \emph{de-dispersing} the routing distribution.

We use $\bar{H} > 0.85 \cdot \log_2 E$ as a practical diagnostic for identifying over-dispersed regimes, not a sharp physical threshold: it is a soft guideline that separates regimes where global importance signals remain reliable (Mixtral at 74.3\%) from those where they degrade (gpt-oss at 90.8\%), with an 11\% margin between our two test architectures. Figure~\ref{fig:entropy-transition} (Appendix) shows that importance signal quality degrades sharply above this band in simulation, matching our observations.

\subsection{MESA: Minimax Expert Score Allocation}
\label{sec:minimax}

Let $\mathbf{R} \in \mathbb{R}^{L \times E}$ denote per-layer normalized router mass scores and $\mathbf{D}_d \in \mathbb{R}^{L \times E}$ denote per-domain router mass for domain $d \in \{1, \ldots, D\}$. Given a target pruning ratio $r$, the minimax algorithm iteratively adjusts importance scores to protect whichever domain is currently worst-affected. Formally, we seek the pruning set that minimizes the worst-case fractional routing loss across domains:
\begin{equation}
\min_{\mathcal{P} \subseteq [L] \times [E],\; |\mathcal{P}| = \lfloor r \cdot L \cdot E \rfloor} \;\max_{d \in \mathcal{D}} \;\frac{\sum_{(l,e) \in \mathcal{P}} \mathbf{D}_d[l,e]}{\sum_{l,e} \mathbf{D}_d[l,e]}
\label{eq:minimax-obj}
\end{equation}

Algorithm~\ref{alg:minimax} is a greedy heuristic solver for this objective, inspired by distributionally robust optimization~\citep{sagawa2020distributionally}, without formal convergence guarantees. In Algorithm~\ref{alg:minimax}, $\mathcal{P}$ denotes the set of experts selected for \emph{removal} (lowest combined scores). Line~4 computes $\ell_d$, the fraction of domain~$d$'s total routing mass contained in $\mathcal{P}$, i.e., the fractional loss each domain would suffer from this pruning decision. Line~6 identifies the domain suffering the highest fractional loss and Line~10 boosts that domain's scores, making its experts less likely to be pruned in subsequent iterations. Lines 6--7 record the best solution found so far, so the algorithm can explore aggressively without losing progress. The decaying boost rate starts with large adjustments and refines gradually. The full pipeline is: (1)~profile per-domain routing mass from calibration data, (2)~run Algorithm~\ref{alg:minimax}, (3)~apply global ranking to select experts for removal.

\begin{algorithm}[t]
\caption{MESA: Minimax Expert Score Allocation}\label{alg:minimax} \small
\begin{algorithmic}[1]
\REQUIRE Router mass $\mathbf{R} \in \mathbb{R}^{L \times E}$, domain scores
$\{\mathbf{D}_d\}_{d=1}^{D}$, ratio $r$, boost $\eta_0 {=} 0.15$, steps $T {=} 20$
\ENSURE Pruning set $\mathcal{P}^*$ (experts to remove)
\STATE $\mathbf{S} \leftarrow \mathbf{R}$; \; $\mathbf{S}^* \leftarrow \mathbf{R}$; \; $\ell^* \leftarrow \infty$
\FOR{$t = 1$ to $T$}
  \STATE $\mathcal{P} \leftarrow \textsc{Allocate}(\mathbf{S}, r)$ \hfill\COMMENT{Global ranking}
  \STATE $\ell_d \leftarrow \sum_{(l,e) \in \mathcal{P}} \mathbf{D}_d[l,e] \;/\; \sum_{l,e} \mathbf{D}_d[l,e]$ \hfill\COMMENT{$\forall d \in \{1,\ldots,D\}$}
  \STATE $d^* \leftarrow \arg\max_d \; \ell_d$ \hfill \COMMENT{Worst domain}
  \IF{$\ell_{d^*} < \ell^*$}
    \STATE $\ell^* \leftarrow \ell_{d^*}$; $\mathbf{S}^* \leftarrow \mathbf{S}$
  \ENDIF
  \STATE $\eta_t \leftarrow \eta_0 \cdot (1 - 0.5 t / (T{-}1))$ \hfill\COMMENT{Decay}
  \STATE $\mathbf{S} \leftarrow \textsc{RowNorm}(\mathbf{S} + \eta_t \cdot \mathbf{D}_{d^*})$ \hfill\COMMENT{Boost + normalize}
\ENDFOR
\RETURN $\mathbf{S}^*$
\end{algorithmic}
\end{algorithm}

\section{Experiments}
\label{sec:experiments}

\subsection{Setup}
\label{sec:setup}

\textbf{Model.}\quad We evaluate on gpt-oss-20b~\citep{openai2025gptoss120bgptoss20bmodel}, a 20B parameter MoE model with 24 transformer layers, 32 experts per layer (768 total), top-4 routing. For cross-architecture validation, we also evaluate on Mixtral-8x7B-Instruct~\citep{jiang2024mixtral} (32 layers, 8 experts, top-2 routing, $\lambda_{\text{aux}} \approx 0.001$).

\textbf{Over-dispersion.}\quad The model was trained with $\lambda_{\text{aux}} = 0.9$~\citep{openai2025gptoss120bgptoss20bmodel}, approximately $900\times$ the default coefficient in Mixtral ($0.001$) and $90\times$ that of Switch Transformer ($0.01$). This produces nearly uniform routing distributions across all 32 experts at every layer.

\textbf{Calibration data.}\quad Importance scoring uses 1000 calibration examples per domain drawn from 6 domains (math, coding, commonsense, science, instruction following, general knowledge). Calibration data consists of topic-level domain text, not benchmark-formatted examples; several of the 11 evaluation benchmarks (e.g., MMLU-Pro, WinoGrande) do not have a dedicated matching calibration domain and are therefore genuinely held out with respect to calibration, so performance on them reflects generalization of the domain-fair scoring beyond the calibrated domains.

\textbf{Evaluation.}\quad We evaluate pruned models on 11 benchmarks spanning math reasoning (GSM8K~\citep{cobbe2021training}, AIME~2025), graduate-level science (GPQA-Diamond~\citep{rein2023gpqagraduatelevelgoogleproofqa}), broad knowledge (MMLU~\citep{hendrycks2021measuring}, MMLU-Pro~\citep{wang2024mmluprorobustchallengingmultitask}), commonsense and science reasoning (ARC-Challenge, ARC-Easy~\citep{clark2018think}, OpenBookQA~\citep{mihaylov2018can}, PIQA~\citep{bisk2020piqa}, WinoGrande~\citep{sakaguchi2021winogrande}), and instruction following (IFEval~\citep{zhou2023instructionfollowingevaluationlargelanguage}, instruction-level loose accuracy). All evaluations use vLLM~\citep{kwon2023efficient} with temperature 0.6, medium reasoning effort, 128K max tokens, and prefix caching enabled. WikiText-2 perplexity is reported separately as a local proxy metric.

\textbf{Baselines.}\quad We compare against four baselines: \textsc{RM-Uniform}~\citep{he2024demystifying} ranks experts by aggregate router mass and prunes the lowest-scoring 25\% uniformly from each layer; REAP~\citep{lasby2026reapexpertspruningprevails} scores experts by activation-weighted routing probability with global budget allocation; EvoESAP~\citep{liu2026evoesapnonuniformexpertpruning} applies evolutionary search for non-uniform per-layer budget allocation using router-mass scores; and random expert pruning serves as a lower bound. All pruned models remove 25\% of experts ($32 \to 24$ per layer).

\subsection{PPL--Accuracy Disconnect}
\label{sec:ppl}

\begin{table}[t]
\centering
\caption{WikiText-2 PPL vs.\ GSM8K accuracy across pruning configurations at $r=0.25$.
PPL and task accuracy are \emph{uncorrelated} in over-dispersed models: random
pruning achieves the lowest PPL (40.95) but near-worst task accuracy, while MESA
\emph{improves} GSM8K above the unpruned baseline despite moderate PPL.}
\label{tab:ppl-disconnect}
\small
\begin{tabular}{@{}lcc@{}}
\toprule
Method & WikiText-2 PPL ($\downarrow$) & GSM8K (\%) ($\uparrow$) \\
\midrule
Baseline (unpruned) & 126.02 & 88.86 \\
MESA (ours) & 107.14 & 92.87 \\
REAP & 145.04 & 89.31 \\
\textsc{RM-Uniform} & 83.29 & 83.24 \\
Random & 40.95 & 78.54 \\
\bottomrule
\end{tabular}
\end{table}

\begin{figure}[t]
\centering
\includegraphics[width=\linewidth]{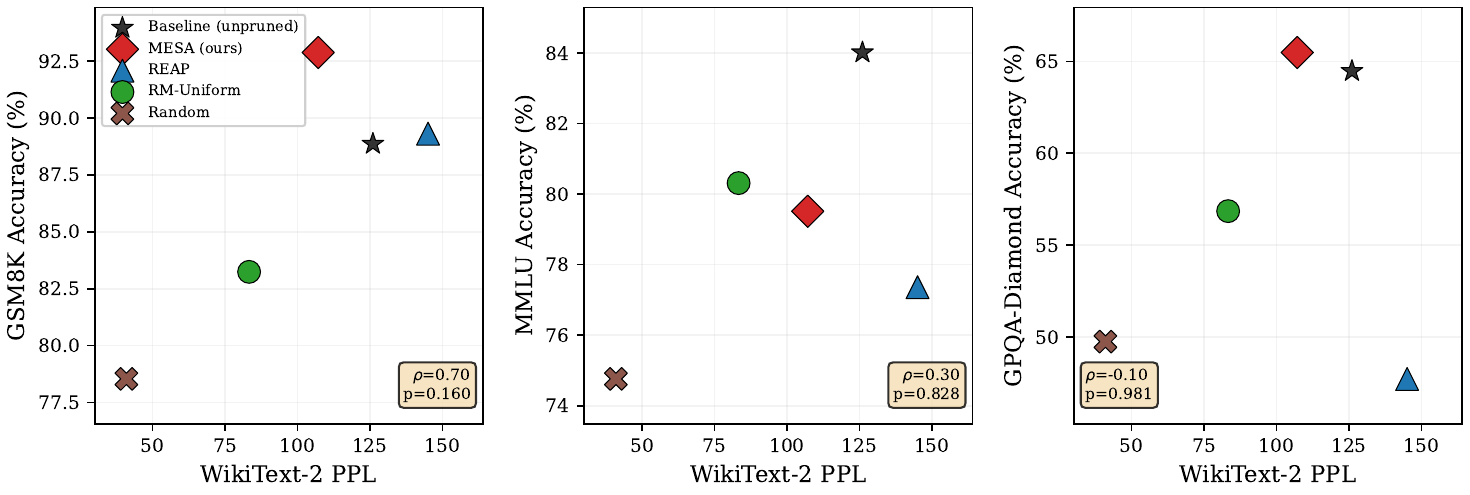}
\caption{PPL vs task accuracy across three benchmarks at $r=0.25$. Each
point is one pruning method. The expected negative correlation (lower PPL $=$ better accuracy) does not hold: $\rho$ is positive on GSM8K and MMLU and near zero on GPQA, confirming PPL is uninformative or misleading for method comparison in over-dispersed models.}
\label{fig:ppl-scatter}
\end{figure}

Table~\ref{tab:ppl-disconnect} illustrates the PPL--accuracy disconnect. Among the methods evaluated on both PPL and task accuracy, REAP achieves the \emph{highest} PPL (145.04, worse than the unpruned baseline of 126.02) yet strong GSM8K accuracy (89.31\%). MESA achieves moderate PPL (107.14) and the best GSM8K among pruned methods (92.87\%), \emph{exceeding} the unpruned baseline (88.86\%). Conversely, random pruning achieves the lowest PPL ($\sim$40), yet degrades task performance severely (AIME'25 33.33\%, GSM8K 78.54\%). This pattern is consistent with the over-dispersion hypothesis: pruning concentrates routing mass onto fewer experts, which can decrease PPL (a token-level metric) while disrupting the model's ability to perform multi-step reasoning. We therefore report only task accuracy for method comparison throughout. Figure~\ref{fig:ppl-scatter} visualizes this disconnect across three representative benchmarks. A detailed explanation of why PPL can decrease under over-dispersed pruning (the top-$k$-then-softmax mechanism) is provided in Appendix~\ref{app:ppl-confound}.

\subsection{Main Results}
\label{sec:main-results}

\begin{table*}[t]
\centering
\caption{Task accuracy (\%) at $r=0.25$ (25\% expert pruning, $32 \to 24$ experts per layer), 5-seed evaluation with 95\% CIs.
Best pruned result per task in \textbf{bold}.
MESA achieves the most balanced profile, with the best or near-best
accuracy on the majority of benchmarks and the smallest worst-case degradation.}
\label{tab:results-r025}
\footnotesize
\setlength{\tabcolsep}{3.5pt}
\resizebox{\textwidth}{!}{%
\begin{tabular}{@{}l ccc cccc cccc@{}}
\toprule
& \multicolumn{3}{c}{Math Reasoning} & \multicolumn{4}{c}{Knowledge \& Commonsense} & \multicolumn{4}{c}{Multi-domain \& Instruction} \\
\cmidrule(lr){2-4} \cmidrule(lr){5-8} \cmidrule(lr){9-12}
Method & AIME'25 & GSM8K & GPQA & ARC-C & ARC-E & OBQA & PIQA & MMLU & MMLU-Pro & WinoGr. & IFEval \\
\midrule
Baseline & 76.67{\tiny$\pm$3.27} & 88.86{\tiny$\pm$0.38} & 64.47{\tiny$\pm$1.46} & 95.30{\tiny$\pm$0.27} & 98.32{\tiny$\pm$0.15} & 92.99{\tiny$\pm$0.45} & 83.34{\tiny$\pm$0.36} & 84.01{\tiny$\pm$0.10} & 72.51{\tiny$\pm$0.22} & 76.78{\tiny$\pm$1.04} & 91.67{\tiny$\pm$0.67} \\
MESA (ours) & 63.33{\tiny$\pm$6.06} & \textbf{92.87}{\tiny$\pm$0.45} & \textbf{65.48}{\tiny$\pm$4.2} & 92.66{\tiny$\pm$0.8} & \textbf{97.60}{\tiny$\pm$0.3}$^*$ & 90.38{\tiny$\pm$1.2} & \textbf{80.73}{\tiny$\pm$1.5} & 79.51{\tiny$\pm$0.6} & \textbf{67.25}{\tiny$\pm$0.9} & 69.59{\tiny$\pm$2.1} & 89.92{\tiny$\pm$1.1} \\
REAP & \textbf{73.33}{\tiny$\pm$9.7} & 89.31{\tiny$\pm$1.8} & 47.72{\tiny$\pm$5.6} & 91.63{\tiny$\pm$1.0} & 95.03{\tiny$\pm$0.7} & \textbf{92.18}{\tiny$\pm$1.0} & 74.90{\tiny$\pm$2.0} & 77.36{\tiny$\pm$0.8} & 64.27{\tiny$\pm$1.2} & \textbf{78.44}{\tiny$\pm$2.4} & \textbf{90.88}{\tiny$\pm$0.9} \\
\textsc{RM-Uni.} & 46.67{\tiny$\pm$8.5} & 83.24{\tiny$\pm$1.5} & 56.85{\tiny$\pm$5.1} & 94.36{\tiny$\pm$0.6} & 97.56{\tiny$\pm$0.4} & 92.18{\tiny$\pm$0.9} & 74.85{\tiny$\pm$1.8} & \textbf{80.31}{\tiny$\pm$0.5} & 66.32{\tiny$\pm$0.8} & 69.27{\tiny$\pm$2.3} & 86.07{\tiny$\pm$1.4} \\
Random & 33.33{\tiny$\pm$7.2} & 78.54{\tiny$\pm$2.3} & 49.75{\tiny$\pm$4.8} & 90.01{\tiny$\pm$1.2} & 96.46{\tiny$\pm$0.5} & 86.97{\tiny$\pm$1.5} & 78.61{\tiny$\pm$1.6} & 74.76{\tiny$\pm$0.9} & 56.06{\tiny$\pm$1.3} & 72.91{\tiny$\pm$2.8} & 89.68{\tiny$\pm$1.2} \\
EvoESAP$^\dagger$ & 56.67{\tiny$\pm$8.0} & 78.32{\tiny$\pm$2.0} & 44.16{\tiny$\pm$4.5} & \textbf{94.96}{\tiny$\pm$0.6} & \textbf{97.64}{\tiny$\pm$0.3}$^*$ & 89.18{\tiny$\pm$1.1} & 80.40{\tiny$\pm$1.5} & 80.15{\tiny$\pm$0.5} & 66.69{\tiny$\pm$0.9} & 65.88{\tiny$\pm$2.5} & 74.07{\tiny$\pm$1.3} \\
\bottomrule
\end{tabular}%
}
\vspace{2pt}
{\scriptsize $^\dagger$EvoESAP evolutionary allocation~\citep{liu2026evoesapnonuniformexpertpruning} applied to router-mass scores; it is a literature baseline (evolutionary allocation), not a component of MESA. $^*$ARC-E: MESA (97.60) and EvoESAP (97.64) are statistically indistinguishable (0.04pp apart, within the reported CIs); both are bolded as an effective tie.}
\end{table*}

Table~\ref{tab:results-r025} presents the main results at $r=0.25$. MESA achieves the most balanced profile across all benchmarks, with no single-task collapse. On math reasoning, MESA achieves 92.87\% on GSM8K ($+4.01$\% above the off-the-shelf baseline; Table~\ref{tab:ppl-disconnect}), consistent with the dispersion hypothesis: pruning concentrates routing onto fewer, more capable experts, improving math accuracy. However, the key finding is the \emph{trade-off} between methods: REAP preserves AIME25 ($73.33 \pm 9.7$\% vs.\ baseline 76.67\%) but has 17.76\% gap on GPQA relative to MESA ($47.72 \pm 5.6$\% vs.\ $65.48 \pm 4.2$\%), losing graduate-level science for math reasoning preservation. \textsc{RM-Uniform} exhibits the reverse: it retains knowledge tasks (MMLU 80.31\%, ARC-C 94.36\%) but suffers large reasoning drops (AIME25 $46.67 \pm 8.5$\%, $-30.00$\%). No single method dominates across all capabilities, a trade-off that emerges under over-dispersed routing.

Across the 11-benchmark evaluation, MESA's minimax objective minimizes the worst-affected domain's degradation rather than average-case accuracy: relative to REAP, MESA's largest advantage is on GPQA ($+$17.8\%, the domain REAP degrades most), while REAP's largest advantage is on AIME25 ($+$10.0\%, the domain MESA degrades most). On a per-benchmark basis, this corresponds to MESA outperforming REAP on 7 tasks (GPQA $+$17.8\%, PIQA $+$5.8\%, GSM8K $+$3.6\%, MMLU-Pro $+$3.0\%, ARC-E $+$2.6\%, MMLU $+$2.2\%, ARC-C $+$1.0\%) while REAP leads on 4 (AIME25 $+$10.0\%, WinoGrande $+$8.9\%, OBQA $+$1.8\%, IFEval $+$1.0\%). The GPQA gap is statistically significant ($z = 3.57$, $p < 0.001$ on 198 questions).

\begin{table*}[t]
\centering
\caption{Per-benchmark accuracy \emph{change} ($\Delta$, percentage points) relative to the
unpruned baseline, gpt-oss-20B at $r{=}0.25$, single-seed evaluation across all 11 benchmarks.
Best (least-negative or most-positive) $\Delta$ per row in \textbf{bold}; ties bolded.}
\label{tab:delta-11}
\footnotesize
\setlength{\tabcolsep}{4pt}
\resizebox{\textwidth}{!}{%
\begin{tabular}{@{}l ccccc@{}}
\toprule
Benchmark & MESA (ours) & REAP & EvoESAP & RM-Uniform & Random \\
\midrule
AIME'25          & $-13.34$          & \textbf{$-3.34$} & $-20.00$ & $-30.00$ & $-43.34$ \\
GSM8K            & \textbf{$+4.01$}  & $+0.45$          & $-10.54$ & $-5.62$  & $-10.32$ \\
ARC-C            & $-2.64$           & $-3.67$          & \textbf{$-0.34$} & $-0.94$ & $-5.29$ \\
ARC-E            & $-0.72$           & $-3.29$          & \textbf{$-0.68$} & $-0.76$ & $-1.86$ \\
GPQA Diamond     & \textbf{$+1.01$}  & $-16.75$         & $-20.31$ & $-7.62$ & $-14.72$ \\
OpenBookQA       & $-2.61$           & \textbf{$-0.81$} & $-3.81$  & \textbf{$-0.81$} & $-6.02$ \\
MMLU             & $-4.50$           & $-6.65$          & $-3.86$  & \textbf{$-3.70$} & $-9.25$ \\
MMLU-Pro         & \textbf{$-5.26$}  & $-8.24$          & $-5.82$  & $-6.19$  & $-16.45$ \\
IFEval           & $-1.75$           & \textbf{$-0.79$} & $-17.60$ & $-5.60$  & $-1.99$ \\
PIQA             & \textbf{$-2.61$}  & $-8.44$          & $-2.94$  & $-8.49$  & $-4.73$ \\
WinoGrande       & $-7.19$           & \textbf{$+1.66$} & $-10.90$ & $-7.51$  & $-3.87$ \\
\bottomrule
\end{tabular}%
}
\vspace{2pt}
\end{table*}

\subsection{Deployment Footprint}
\label{sec:footprint}

Because expert weights dominate an MoE's parameter count, removing a quarter of the experts at $r=0.25$ ($32 \to 24$ per layer) reduces the served model's parameter count, and hence its resident memory footprint, by approximately the same fraction. This is the practical motivation for expert pruning: the entire expert population must be resident even though each token activates only top-$k$ of them, so total parameter count rather than per-token compute sets the memory and GPU budget.

We note explicitly what pruning at this ratio does \emph{not} buy. Removing experts from the pool leaves top-$k$ unchanged, so each token still routes through $k$ expert FFNs and per-request decode latency is essentially unaffected; on gpt-oss-20b the MXFP4 expert weights are small enough ($\sim$1.6\,MB per expert per GPU) that they are not the binding constraint on single-request throughput. Reducing per-token compute requires changing $k$ or the expert internals, which is orthogonal to the allocation question studied here. We therefore frame MESA's contribution as domain-fair accuracy retention at a reduced memory footprint, and make no per-request speedup claim.

\subsection{Cross-Architecture Validation}
\label{sec:crossarch}

To test whether MESA's benefit is specific to over-dispersed models, we evaluate on two contrasting architectures: Mixtral-8x7B-Instruct~\citep{jiang2024mixtral} (8 experts, top-2 routing, $\lambda_{\text{aux}} \approx 0.001$), which has standard (non-dispersed) routing, and OLMoE-1B-7B-0924 (16 layers, 64 experts, top-8), an independently trained, fully-open \emph{base} model that is itself over-dispersed ($\bar{H}=0.925$). The regime, not the architecture, predicts whether MESA helps: it does not help on Mixtral, and it does on OLMoE (Section~\ref{sec:olmoe}).

On Mixtral, PPL behaves \emph{normally}: pruning increases perplexity (baseline 4.78 $\to$ REAP 6.00, Random 8.31, RM-Uniform 14.35, MESA 16.43; Table~\ref{tab:ppl-mixtral}), the standard expected behavior. This differs from gpt-oss, where pruning \emph{decreases} PPL (baseline 126.02 $\to$ Random 40.95). MESA also causes the largest PPL degradation on Mixtral ($+244\%$) despite being competitive on gpt-oss, while REAP achieves the best PPL (6.00, $+26\%$ from baseline).

\begin{table}[t]
\centering
\caption{Cross-architecture validation on Mixtral-8x7B-Instruct ($\lambda_{\text{aux}} \approx 0.001$, $r=0.25$, $8 \to 6$ experts).
Unlike gpt-oss, Mixtral shows PPL \emph{increases} with pruning (standard behavior).
REAP achieves the best PPL (6.00), while MESA, designed for over-dispersed routing, produces the worst (16.43).}
\label{tab:ppl-mixtral}
\small
\begin{tabular}{@{}lc@{}}
\toprule
Method & WikiText-2 PPL \\
\midrule
Baseline (8E) & 4.78 \\
REAP (6E) & \textbf{6.00} \\
Random (6E) & 8.31 \\
\textsc{RM-Uniform} (6E) & 14.35 \\
MESA (6E) & 16.43 \\
\bottomrule
\end{tabular}
\end{table}

\begin{table}[t]
\centering
\caption{Cross-architecture task accuracy (\%) on Mixtral-8x7B-Instruct at $r=0.25$ ($8 \to 6$ experts),
5-seed means $\pm$ std. Best pruned result per task in \textbf{bold}. REAP retains 98\% of baseline
average accuracy while MESA retains 63\%.}
\label{tab:crossarch}
\footnotesize
\setlength{\tabcolsep}{5pt}
\begin{tabular}{@{}l ccccccc@{}}
\toprule
Method & ARC-C & ARC-E & GPQA & GSM8K & MMLU & IFEval & PIQA \\
\midrule
Baseline & 85.29{\tiny$\pm$0.24} & 94.51{\tiny$\pm$0.15} & 31.07{\tiny$\pm$1.21} & 31.42{\tiny$\pm$1.25} & 68.54{\tiny$\pm$0.16} & 66.58{\tiny$\pm$1.34} & 85.60{\tiny$\pm$0.34} \\
\midrule
REAP & \textbf{82.19}{\tiny$\pm$0.30} & \textbf{92.29}{\tiny$\pm$0.15} & 28.93{\tiny$\pm$2.61} & 35.00{\tiny$\pm$0.97} & \textbf{64.29}{\tiny$\pm$0.19} & 65.23{\tiny$\pm$0.73} & \textbf{83.58}{\tiny$\pm$0.26} \\
EvoESAP & 73.63{\tiny$\pm$0.32} & 86.14{\tiny$\pm$0.30} & 31.68{\tiny$\pm$1.59} & \textbf{40.56}{\tiny$\pm$0.66} & 56.65{\tiny$\pm$0.27} & \textbf{65.62}{\tiny$\pm$0.23} & 77.21{\tiny$\pm$0.38} \\
Random & 70.13{\tiny$\pm$0.50} & 80.61{\tiny$\pm$0.29} & 19.70{\tiny$\pm$1.36} & 20.71{\tiny$\pm$0.84} & 52.56{\tiny$\pm$0.11} & 57.98{\tiny$\pm$1.25} & 74.95{\tiny$\pm$0.46} \\
MESA & 51.97{\tiny$\pm$1.20} & 67.15{\tiny$\pm$1.13} & 22.34{\tiny$\pm$5.74} & 3.49{\tiny$\pm$0.48} & 38.49{\tiny$\pm$0.34} & 49.44{\tiny$\pm$1.51} & 58.42{\tiny$\pm$0.71} \\
\textsc{RM-Uniform} & 49.77{\tiny$\pm$1.05} & 60.23{\tiny$\pm$0.60} & \textbf{37.66}{\tiny$\pm$0.22} & 2.96{\tiny$\pm$0.38} & 24.60{\tiny$\pm$2.36} & 45.95{\tiny$\pm$1.68} & 57.04{\tiny$\pm$1.14} \\
\bottomrule
\end{tabular}
\end{table}

Under standard routing, experts are specialized: MESA's minimax objective interprets this specialization as unfair utilization and prunes critical domain experts. REAP's activation-weighted score remains informative across both regimes, making it regime-agnostic. EvoESAP shows that evolutionary search can find beneficial pruning configurations even under standard routing (GSM8K exceeds baseline by +9.14\%), suggesting that non-uniform allocation adapts to the routing structure regardless of regime.

Table~\ref{tab:crossarch} supports the regime-specificity hypothesis with task accuracy data. On Mixtral, REAP retains 98\% of baseline average accuracy and EvoESAP retains 93\%, improving GSM8K above baseline ($+9.14$\%). MESA, the method that preserves GPQA on over-dispersed gpt-oss, retains 63\%, indicating that domain-fair scoring addresses a failure specific to over-dispersed routing.

\subsection{Generality to a Third Over-Dispersed Family: OLMoE-1B-7B}
\label{sec:olmoe}

The over-dispersed regime characterized above is not unique to gpt-oss or to instruction-tuned
models. We test a third, independently-trained MoE family: \textbf{OLMoE-1B-7B-0924}, a fully-open
\emph{base} (not instruction-tuned) model from a different lab with 16 layers, 64 experts, and top-8
routing. Its native per-token routing entropy is $\bar{H}=0.925\cdot\log_2 E$, placing it firmly in the
over-dispersed regime by our practical diagnostic ($\bar{H}>0.85$, a soft guideline rather than a
sharp threshold; Section~\ref{sec:overdispersed}). This point
extends the generality claim across both \emph{base} and \emph{instruction-tuned} training and across a
third architecture.

\begin{table}[t]
\centering
\caption{Generality to a third, independently-trained over-dispersed MoE family: OLMoE-1B-7B-0924
(a fully-open \emph{base} model; 16 layers, 64 experts, top-8; native routing entropy
$\bar{H}=0.925\cdot\log_2 E$). Loglikelihood/cloze accuracy (\%) at $r=0.25$ (25\% expert pruning, $64\to48$
experts per layer), single seed. Best pruned result per task in \textbf{bold}. On these six
commonsense tasks (cloze/acc\_norm), MESA calibrated in the matching \emph{prefill}
regime leads every pruned method and beats REAP on all six, retaining 92.8\% of unpruned commonsense
accuracy.}
\label{tab:olmoe}
\footnotesize
\setlength{\tabcolsep}{4pt}
\begin{tabular}{@{}l ccccc@{}}
\toprule
Task (cloze acc\_norm) & Unpruned & \textbf{MESA (prefill)} & MESA (gen) & REAP (gen) & \textsc{RM} (gen) \\
\midrule
OpenBookQA        & 45.2 & \textbf{39.8} & 35.8 & 38.4 & 35.8 \\
PIQA              & 80.6 & \textbf{77.6} & 74.2 & 71.9 & 72.2 \\
WinoGrande        & 72.0 & \textbf{67.7} & 59.0 & 67.6 & 59.0 \\
ARC-Challenge     & 55.2 & \textbf{50.0} & 44.0 & 40.6 & 45.8 \\
ARC-Easy          & 82.3 & \textbf{77.4} & 72.2 & 64.0 & 73.2 \\
HellaSwag         & 76.0 & \textbf{69.2} & 58.5 & 64.9 & 57.7 \\
\midrule
\textbf{Commonsense avg} & \textbf{68.6} & \textbf{63.6} & 57.3 & 57.9 & 57.3 \\
\% of unpruned    & 100\% & \textbf{92.8\%} & 83.5\% & 84.4\% & 83.5\% \\
\bottomrule
\end{tabular}
\vspace{2pt}

{\scriptsize The gen-route baselines are calibrated under decode-time routing, while MESA (prefill)
is calibrated under a single forward pass; a fully prefill-vs-prefill comparison across all methods,
together with multi-seed replication, is left to future work.}
\end{table}

\begin{table}[t]
\centering
\caption{OLMoE-1B-7B prefill-MESA ratio sweep (single seed): commonsense accuracy degrades
gracefully with no cliff. Eval-matched (prefill) calibration buys $\sim$one extra ratio step of
headroom: prefill MESA at $r=0.375$ (57.2) matches every gen-route method at $r=0.25$ ($57.3$--$57.9$;
Table~\ref{tab:olmoe}).}
\label{tab:olmoe-sweep}
\footnotesize
\setlength{\tabcolsep}{5pt}
\begin{tabular}{@{}l cccc@{}}
\toprule
Task & Unpruned & MESA $r{=}0.25$ & MESA $r{=}0.375$ & MESA $r{=}0.50$ \\
\midrule
OpenBookQA    & 45.2 & 39.8 & 35.6 & 31.2 \\
PIQA          & 80.6 & 77.6 & 72.4 & 68.1 \\
WinoGrande    & 72.0 & 67.7 & 58.6 & 53.8 \\
ARC-Challenge & 55.2 & 50.0 & 43.5 & 34.5 \\
ARC-Easy      & 82.3 & 77.4 & 73.4 & 53.7 \\
HellaSwag     & 76.0 & 69.2 & 59.6 & 46.2 \\
\midrule
\textbf{Commonsense avg} & \textbf{68.6} & \textbf{63.6} & \textbf{57.2} & \textbf{47.9} \\
\% of unpruned & 100\% & \textbf{92.8\%} & 83.4\% & 69.9\% \\
\bottomrule
\end{tabular}
\end{table}

\textbf{Capture-regime $\leftrightarrow$ eval-regime matching.}\quad
OLMoE surfaced a mechanism the other models did not, because they were scored generatively while OLMoE,
a base model, is scored with loglikelihood/cloze multiple choice (a single teacher-forced forward
pass with \emph{no} generation). When we calibrated routing with the decode-time (gen-route) harvest that
wins on gpt-oss-20b ($+8.5$pp) and Mixtral ($+13$pp), MESA and \textsc{RM-Uniform} produced nearly
identical pruned sets. Under decode the four calibration domains routed almost identically (per-domain
routing-mass correlation $0.57$--$0.86$), so the minimax objective had no per-domain signal to grip and
degenerated toward router mass. Calibrating in the \emph{same} regime the benchmark scores in
(\textbf{prefill}: one forward pass over the calibration text, no generation) restored per-domain
structure: cross-domain routing-mass correlation collapsed to $\approx 0$ (math--knowledge $-0.068$,
science--knowledge $0.004$, instruction--knowledge $-0.066$), and the overlap between MESA's and
\textsc{RM-Uniform}'s pruned sets fell to Jaccard $0.705$, the lowest we observe, versus
$\approx 1.0$ under decode (Jaccard of MESA vs.\ REAP was $0.252$). We report this as a
\emph{correlational/observational} pattern rather than a causal claim: when routing is calibrated in
the regime the benchmark scores in, minimax-fair recovers per-domain grip and the pruned set protects
the experts the eval actually exercises.

\textbf{Headline.}\quad
Under eval-matched (prefill) calibration, MESA is the best pruned method on all six commonsense tasks and
beats REAP on every one of them (Table~\ref{tab:olmoe}): commonsense average $63.6$ vs.\ $57$--$58$ for
every gen-route method (a $\sim$6pp gap), retaining \textbf{92.8\% of unpruned commonsense accuracy}
at $r=0.25$. The gain is consistent with, not a contradiction of, the over-dispersed-regime prediction
that domain-aware scoring protects whichever domains a global metric would wash out. Degradation is
graceful across ratios ($92.8\% \to 83.4\% \to 69.9\%$ at $r=0.25/0.375/0.50$; Table~\ref{tab:olmoe-sweep}),
and eval-matched calibration buys roughly \emph{one extra ratio step} of headroom: prefill MESA at
$r=0.375$ ($57.2$) matches every gen-route method at $r=0.25$ ($57.3$--$57.9$).

\textbf{Compute.}\quad
The full OLMoE pipeline is static $L\times E$ score matrices plus one \emph{prefill} forward pass per
calibration prompt and the apply step, with \emph{zero} eval-time forward passes for scoring, matching the
compute profile of Algorithm~\ref{alg:minimax} on the other models.

\textbf{Limitations.}\quad
All OLMoE results are single measurements; the $r=0.25$ leads are large ($>5$pp) relative to typical
seed-to-seed variance on these benchmarks, but multi-seed replication is left to future work. The
reported REAP/\textsc{RM}/gen-route-MESA baselines are calibrated under decode-time routing while the
headline MESA result is calibrated under prefill; a fully prefill-vs-prefill comparison across all
methods (REAP's plan is domain-independent so it is unaffected; the unpruned row is
regime-independent) would strengthen the comparison further.

\subsection{Generality to a Third Architecture: Gemma-4-26B-A4B-it}
\label{sec:generality-gemma}

We extend the generality evidence to \textbf{Gemma-4-26B-A4B-it}, a third distinct architecture (128
routed experts, top-8 routing) at $r=0.25$, evaluated under two protocols: temperature $0.6$ (main
text) and temperature $0$, deterministic decoding, as an appendix replication. (We additionally test
scale and pruning-ratio severity within the gpt-oss family itself at $r=0.5$ on gpt-oss-120B; see
Appendix~\ref{app:gptoss120b}.)

\textbf{Gemma-4-26B-A4B-it: the flagship cross-architecture result.}\quad
Table~\ref{tab:results-gemma} reports all 11 benchmarks across MESA, REAP, \textsc{RM-Uniform}, and
the unpruned model. MESA matches or exceeds both baselines on 10 of the 11 benchmarks (the sole
exception, ARC-Challenge, is a $0.29$pp gap to REAP within single-measurement noise), with the
widest margins on AIME'25 ($36.7$pp over REAP) and GPQA-Diamond, the hardest task in the suite, where MESA leads REAP by $14.3$pp
($67.85$ vs.\ $53.54$) and \textsc{RM-Uniform} by $21.2$pp ($67.85$ vs.\ $46.63$). This is the
cleanest cross-architecture evidence for worst-domain protection reported in this paper: on an
architecture with a different expert count, routing width, and training lineage than either gpt-oss
model, the minimax objective still identifies and protects the domain that would otherwise degrade
most.

\begin{table*}[t]
\centering
\caption{Task accuracy (\%) on \textbf{Gemma-4-26B-A4B-it at $r=0.25$} (128 routed experts,
top-8; $128 \to 96$ experts per layer), temperature $0.6$, single measurement per cell.
Best pruned result per task in \textbf{bold}. MESA matches or exceeds both REAP and
\textsc{RM-Uniform} on every one of the 11 benchmarks, with the largest gap on GPQA-Diamond, the
hardest task in the suite ($+14.3$pp over REAP, $+21.2$pp over \textsc{RM-Uniform}).}
\label{tab:results-gemma}
\footnotesize
\setlength{\tabcolsep}{3.5pt}
\resizebox{\textwidth}{!}{%
\begin{tabular}{@{}l ccc cccc cccc@{}}
\toprule
& \multicolumn{3}{c}{Math Reasoning} & \multicolumn{4}{c}{Knowledge \& Commonsense} & \multicolumn{4}{c}{Multi-domain \& Instruction} \\
\cmidrule(lr){2-4} \cmidrule(lr){5-8} \cmidrule(lr){9-12}
Method & AIME'25 & GSM8K & GPQA & ARC-C & ARC-E & OBQA & PIQA & MMLU & MMLU-Pro & WinoGr. & IFEval \\
\midrule
Unpruned & 75.56 & 96.41 & 71.21 & 96.62 & 98.78 & 94.47 & 90.82 & 86.58 & 82.05 & 80.27 & 89.81 \\
MESA (ours) & \textbf{72.22} & \textbf{96.21} & \textbf{67.85} & 95.53 & \textbf{98.30} & \textbf{92.13} & \textbf{88.03} & \textbf{83.73} & \textbf{79.40} & \textbf{77.56} & \textbf{88.98} \\
REAP & 35.56 & 95.60 & 53.54 & \textbf{95.82} & 98.22 & 90.67 & 86.85 & 78.94 & 67.90 & 76.95 & 88.70 \\
\textsc{RM-Uni.} & 50.00 & 95.12 & 46.63 & 91.52 & 96.55 & 88.13 & 81.27 & 78.26 & 69.56 & 73.24 & 86.67 \\
\bottomrule
\end{tabular}%
}
\vspace{2pt}
{\scriptsize ARC-Challenge is a nominal exception (REAP $95.82$ vs.\ MESA $95.53$, a $0.29$pp gap within
single-measurement noise); MESA is tied-or-better on all other tasks. Single measurement per cell,
temperature $0.6$; see Appendix Table~\ref{tab:gemma-temp0} for an independent temperature-$0$
(deterministic, 3-seed) replication of the worst-domain result on a 5-task subset.}
\end{table*}

\textbf{Replication and mechanism check.}\quad
Appendix Table~\ref{tab:gemma-temp0} replicates the Gemma-4 worst-domain result under deterministic
(temperature-$0$), 3-seed decoding, with non-overlapping confidence intervals on GPQA-Diamond
($65.66\pm1.82$ MESA vs.\ $55.56\pm1.95$ REAP), confirming the temperature-$0.6$ finding is not an
artifact of sampling temperature. A domain-aware REAP control -- REAP given the same per-domain
conditioning as MESA -- reaches GPQA-Diamond only $58.4$, well short of MESA's $65.66$--$67.85$,
indicating that MESA's gain comes from the minimax worst-case \emph{objective} itself rather than
merely from conditioning the importance score by domain.

\textbf{Regime status.}\quad
Gemma-4-26B-A4B-it is \emph{not} over-dispersed by our diagnostic (native per-token routing entropy
$\bar{H}\approx0.69$); we present it solely as evidence that MESA's worst-domain protection
generalizes to an architecturally distinct, concentrated-routing model, not as an additional
over-dispersed-regime data point. Appendix~\ref{app:gptoss120b} presents a complementary,
same-family scale/ratio generality check (gpt-oss-120B at $r=0.5$).

\section{Analysis}
\label{sec:analysis}

\subsection{Capability Trade-off}
\label{sec:frontier}

The task-dependent scoring effect observed above, where REAP excels at generation tasks but loses accuracy on knowledge-intensive MC tasks while RM-Uniform shows the reverse, decomposes along two axes. PCA on the accuracy matrix identifies two principal components: PC1 loads on reasoning/generative tasks (GSM8K, AIME'25), while PC2 loads on knowledge/pattern-matching tasks (MMLU, ARC). Method rankings invert across these clusters (permutation test $p = 0.0004$; see Appendix~\ref{app:permtest}), indicating a structured capability trade-off.

\begin{table}[t]
\centering
\caption{Reasoning/Knowledge degradation at $r=0.25$ (single-seed results).
$\Delta_{\text{R}}$: mean accuracy change on AIME'25 and GSM8K;
$\Delta_{\text{K}}$: mean change on ARC-C and ARC-E.
Each method occupies a distinct position on the capability trade-off; MESA
achieves the most balanced trade-off with a net reasoning \emph{improvement}
on GSM8K offsetting its AIME'25 drop.}
\label{tab:rk-ratio}
\small
\begin{tabular}{@{}lccc@{}}
\toprule
Method & $\Delta_{\text{R}}$ (\%) & $\Delta_{\text{K}}$ (\%) & Character \\
\midrule
MESA (ours) & $-4.67$ & $-1.68$ & Balanced \\
REAP & $-1.45$ & $-3.48$ & Reasoning-preserving \\
EvoESAP$^\dagger$ & $-15.27$ & $-0.51$ & Knowledge-preserving \\
\textsc{RM-Uniform} & $-17.81$ & $-0.85$ & Knowledge-preserving \\
Random & $-26.83$ & $-3.58$ & Severe collapse \\
\bottomrule
\end{tabular}
\end{table}

\begin{figure}[t]
\centering
\includegraphics[width=0.6\linewidth]{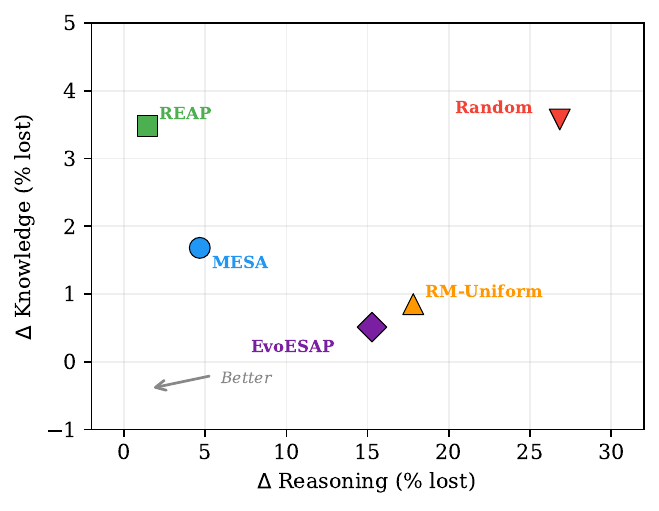}
\caption{Capability trade-off: reasoning vs knowledge preservation at
$r=0.25$. Each method occupies a distinct position; MESA is closest to the
origin (balanced), while REAP favors reasoning preservation.}
\label{fig:pareto}
\end{figure}

Table~\ref{tab:rk-ratio} shows a clear asymmetry. RM-Uniform and Random severely degrade reasoning ($\Delta_{\text{R}} < -17$\% on AIME'25 + GSM8K), while REAP preserves reasoning ($\Delta_{\text{R}} = -1.45$\%) at the cost of knowledge tasks ($\Delta_{\text{K}} = -3.48$\%). MESA achieves modest reasoning degradation ($-4.67$\%, dominated by the AIME'25 drop; GSM8K actually \emph{improves} $+4.01$\% above baseline) while keeping knowledge degradation small ($-1.68$\%), exhibiting the most balanced profile. No single method dominates: the choice depends on which capabilities matter for the deployment. MESA occupies the balanced position nearest the origin on both axes (Figure~\ref{fig:pareto}), appropriate when neither reasoning nor knowledge can be sacrificed disproportionately. A qualitative example of knowledge degradation (REAP losing a specific physics concept on GPQA while retaining the general technique) is provided in Appendix~\ref{app:capability-cliff}.

\subsection{Calibration Efficiency}
\label{sec:calibration}

MESA converges at $s{=}1000$ samples, $l{=}8192$ tokens (97.9\% prune map agreement with the reference configuration). Even at $s{=}20$ samples per domain, MESA achieves 96.4\% agreement, making it practical with minimal calibration data. Sequence length dominates sample count as the primary driver of stability (+15.0\% from doubling sequence length vs.\ +5.8\% from $10\times$ more samples). Compared to REAP's calibration requirement (24{,}576 samples at 16{,}384 tokens, $402.7$M tokens~\citep{lasby2026reapexpertspruningprevails}), MESA's $1000$ samples at $8192$ tokens across 6 domains ($49.2$M tokens) is approximately $8\times$ less calibration compute. Full ablation details are in Appendix~\ref{app:ablations}.

\begin{table}[t]
\centering
\caption{Qualitative compute characterization of expert-scoring and allocation for MESA (ours) vs.\
REAP and Router-Mass (RM). MESA's allocation step (Algorithm~\ref{alg:minimax}) operates entirely on a
small, pre-computed $L\times E$ router-mass matrix (e.g., $24 \times 32$ for gpt-oss-20B) rather than
requiring additional model forward passes, making it structurally cheaper than activation-based scoring
methods such as REAP. No wall-clock timings are claimed here; see \S\ref{sec:calibration} for the
verified $\sim$49$\times$ calibration-sample-count reduction relative to REAP.}
\label{tab:alg-compute}
\small
\begin{tabular}{@{}l p{3.1cm} p{3.1cm} p{2.6cm}@{}}
\toprule
Method & Scoring compute & Allocation compute & Extra fwd.\ passes beyond harvest \\
\midrule
MESA (ours) & Static per-domain $L\times E$ router-mass matrix, already produced by the calibration harvest & CPU-only iterative minimax reweighting over the $L\times E$ matrix ($\sim$15--20 steps) & 0 \\
REAP~\citep{lasby2026reapexpertspruningprevails} & Per-expert activation-norm scoring requires a dedicated forward pass over its calibration set (24{,}576 samples at 16{,}384 tokens) & Global budget allocation over per-expert activation-norm scores & Full calibration-set forward pass (activation-norm scoring) \\
RM & Router-mass matrix only, identical byproduct of the same harvest MESA consumes & Uniform per-layer threshold on aggregate router mass & 0 \\
\bottomrule
\end{tabular}
\end{table}

\section{Conclusion}
\label{sec:conclusion}

Over-dispersed MoE routing, induced by aggressive load-balancing ($\lambda_\text{aux} = 0.9$), constitutes a qualitatively distinct pruning regime: perplexity becomes unreliable as a quality signal, and a capability trade-off emerges in which no single scoring metric dominates across tasks. Cross-architecture experiments confirm this is regime-specific: under standard routing (Mixtral, $\lambda_\text{aux} \approx 0.001$), all methods degrade proportionally and no domain imbalance emerges. MESA validates the practical relevance of this characterization by minimizing the worst-affected domain's degradation rather than average-case accuracy, which on a per-benchmark basis corresponds to outperforming REAP on 7 of 11 benchmarks at $r{=}0.25$, though it does not survive aggressive pruning ($r{=}0.50$), where REAP's activation-aware signal remains robust.

\textbf{Limitations.}\quad Our evaluation covers two architectures (gpt-oss-20b and Mixtral-8x7B-Instruct); generalization to other over-dispersed models remains to be verified. Algorithm~\ref{alg:minimax} is a greedy heuristic without formal convergence guarantees, though empirical ablations show stable convergence within 15 iterations. We evaluate at a single pruning ratio ($r=0.25$) in the main results; behavior at more aggressive ratios requires further study. Future work should explore adaptive methods that detect the routing regime automatically and select scoring accordingly.

{\small
\bibliographystyle{plainnat}
\bibliography{bibliography}

@misc{lasby2026reapexpertspruningprevails,
      title={REAP the Experts: Why Pruning Prevails for One-Shot MoE compression}, 
      author={Mike Lasby and Ivan Lazarevich and Nish Sinnadurai and Sean Lie and Yani Ioannou and Vithursan Thangarasa},
      year={2026},
      eprint={2510.13999},
      archivePrefix={arXiv},
      primaryClass={cs.LG},
      url={https://arxiv.org/abs/2510.13999}, 
}

@misc{openai2025gptoss120bgptoss20bmodel,
      title={gpt-oss-120b \& gpt-oss-20b Model Card}, 
      author={OpenAI and : and Sandhini Agarwal and Lama Ahmad and Jason Ai and Sam Altman and Andy Applebaum and Edwin Arbus and Rahul K. Arora and Yu Bai and Bowen Baker and Haiming Bao and Boaz Barak and Ally Bennett and Tyler Bertao and Nivedita Brett and Eugene Brevdo and Greg Brockman and Sebastien Bubeck and Che Chang and Kai Chen and Mark Chen and Enoch Cheung and Aidan Clark and Dan Cook and Marat Dukhan and Casey Dvorak and Kevin Fives and Vlad Fomenko and Timur Garipov and Kristian Georgiev and Mia Glaese and Tarun Gogineni and Adam Goucher and Lukas Gross and Katia Gil Guzman and John Hallman and Jackie Hehir and Johannes Heidecke and Alec Helyar and Haitang Hu and Romain Huet and Jacob Huh and Saachi Jain and Zach Johnson and Chris Koch and Irina Kofman and Dominik Kundel and Jason Kwon and Volodymyr Kyrylov and Elaine Ya Le and Guillaume Leclerc and James Park Lennon and Scott Lessans and Mario Lezcano-Casado and Yuanzhi Li and Zhuohan Li and Ji Lin and Jordan Liss and Lily and Liu and Jiancheng Liu and Kevin Lu and Chris Lu and Zoran Martinovic and Lindsay McCallum and Josh McGrath and Scott McKinney and Aidan McLaughlin and Song Mei and Steve Mostovoy and Tong Mu and Gideon Myles and Alexander Neitz and Alex Nichol and Jakub Pachocki and Alex Paino and Dana Palmie and Ashley Pantuliano and Giambattista Parascandolo and Jongsoo Park and Leher Pathak and Carolina Paz and Ludovic Peran and Dmitry Pimenov and Michelle Pokrass and Elizabeth Proehl and Huida Qiu and Gaby Raila and Filippo Raso and Hongyu Ren and Kimmy Richardson and David Robinson and Bob Rotsted and Hadi Salman and Suvansh Sanjeev and Max Schwarzer and D. Sculley and Harshit Sikchi and Kendal Simon and Karan Singhal and Yang Song and Dane Stuckey and Zhiqing Sun and Philippe Tillet and Sam Toizer and Foivos Tsimpourlas and Nikhil Vyas and Eric Wallace and Xin Wang and Miles Wang and Olivia Watkins and Kevin Weil and Amy Wendling and Kevin Whinnery and Cedric Whitney and Hannah Wong and Lin Yang and Yu Yang and Michihiro Yasunaga and Kristen Ying and Wojciech Zaremba and Wenting Zhan and Cyril Zhang and Brian Zhang and Eddie Zhang and Shengjia Zhao},
      year={2025},
      eprint={2508.10925},
      archivePrefix={arXiv},
      primaryClass={cs.CL},
      url={https://arxiv.org/abs/2508.10925}, 
}

@misc{shazeer2017outrageously,
      title={Outrageously Large Neural Networks: The Sparsely-Gated Mixture-of-Experts Layer}, 
      author={Noam Shazeer and Azalia Mirhoseini and Krzysztof Maziarz and Andy Davis and Quoc Le and Geoffrey Hinton and Jeff Dean},
      year={2017},
      eprint={1701.06538},
      archivePrefix={arXiv},
      primaryClass={cs.LG},
      url={https://arxiv.org/abs/1701.06538}, 
}

@misc{fedus2022switch,
      title={Switch Transformers: Scaling to Trillion Parameter Models with Simple and Efficient Sparsity}, 
      author={William Fedus and Barret Zoph and Noam Shazeer},
      year={2022},
      eprint={2101.03961},
      archivePrefix={arXiv},
      primaryClass={cs.LG},
      url={https://arxiv.org/abs/2101.03961}, 
}

@misc{lu2024not,
      title={Not All Experts are Equal: Efficient Expert Pruning and Skipping for Mixture-of-Experts Large Language Models}, 
      author={Xudong Lu and Qi Liu and Yuhui Xu and Aojun Zhou and Siyuan Huang and Bo Zhang and Junchi Yan and Hongsheng Li},
      year={2024},
      eprint={2402.14800},
      archivePrefix={arXiv},
      primaryClass={cs.CL},
      url={https://arxiv.org/abs/2402.14800}, 
}

@misc{he2024demystifying,
      title={Towards Efficient Mixture of Experts: A Holistic Study of Compression Techniques}, 
      author={Shwai He and Daize Dong and Liang Ding and Ang Li},
      year={2025},
      eprint={2406.02500},
      archivePrefix={arXiv},
      primaryClass={cs.LG},
      url={https://arxiv.org/abs/2406.02500}, 
}

@misc{chen2025retrainingfreemergingsparsemoe,
      title={Retraining-Free Merging of Sparse MoE via Hierarchical Clustering}, 
      author={I-Chun Chen and Hsu-Shen Liu and Wei-Fang Sun and Chen-Hao Chao and Yen-Chang Hsu and Chun-Yi Lee},
      year={2025},
      eprint={2410.08589},
      archivePrefix={arXiv},
      primaryClass={cs.LG},
      url={https://arxiv.org/abs/2410.08589}, 
}

@misc{liu2026evoesapnonuniformexpertpruning,
      title={EvoESAP: Non-Uniform Expert Pruning for Sparse MoE}, 
      author={Zongfang Liu and Shengkun Tang and Boyang Sun and Zhiqiang Shen and Xin Yuan},
      year={2026},
      eprint={2603.06003},
      archivePrefix={arXiv},
      primaryClass={cs.LG},
      url={https://arxiv.org/abs/2603.06003}, 
}

@misc{jiang2024mixtral,
      title={Mixtral of Experts}, 
      author={Albert Q. Jiang and Alexandre Sablayrolles and Antoine Roux and Arthur Mensch and Blanche Savary and Chris Bamford and Devendra Singh Chaplot and Diego de las Casas and Emma Bou Hanna and Florian Bressand and Gianna Lengyel and Guillaume Bour and Guillaume Lample and Lélio Renard Lavaud and Lucile Saulnier and Marie-Anne Lachaux and Pierre Stock and Sandeep Subramanian and Sophia Yang and Szymon Antoniak and Teven Le Scao and Théophile Gervet and Thibaut Lavril and Thomas Wang and Timothée Lacroix and William El Sayed},
      year={2024},
      eprint={2401.04088},
      archivePrefix={arXiv},
      primaryClass={cs.LG},
      url={https://arxiv.org/abs/2401.04088}, 
}

@misc{sagawa2020distributionally,
      title={Distributionally Robust Neural Networks for Group Shifts: On the Importance of Regularization for Worst-Case Generalization}, 
      author={Shiori Sagawa and Pang Wei Koh and Tatsunori B. Hashimoto and Percy Liang},
      year={2020},
      eprint={1911.08731},
      archivePrefix={arXiv},
      primaryClass={cs.LG},
      url={https://arxiv.org/abs/1911.08731}, 
}

@misc{bisk2020piqa,
      title={PIQA: Reasoning about Physical Commonsense in Natural Language}, 
      author={Yonatan Bisk and Rowan Zellers and Ronan Le Bras and Jianfeng Gao and Yejin Choi},
      year={2019},
      eprint={1911.11641},
      archivePrefix={arXiv},
      primaryClass={cs.CL},
      url={https://arxiv.org/abs/1911.11641}, 
}

@misc{clark2018think,
      title={Think you have Solved Question Answering? Try ARC, the AI2 Reasoning Challenge}, 
      author={Peter Clark and Isaac Cowhey and Oren Etzioni and Tushar Khot and Ashish Sabharwal and Carissa Schoenick and Oyvind Tafjord},
      year={2018},
      eprint={1803.05457},
      archivePrefix={arXiv},
      primaryClass={cs.AI},
      url={https://arxiv.org/abs/1803.05457}, 
}

@misc{cobbe2021training,
      title={Training Verifiers to Solve Math Word Problems}, 
      author={Karl Cobbe and Vineet Kosaraju and Mohammad Bavarian and Mark Chen and Heewoo Jun and Lukasz Kaiser and Matthias Plappert and Jerry Tworek and Jacob Hilton and Reiichiro Nakano and Christopher Hesse and John Schulman},
      year={2021},
      eprint={2110.14168},
      archivePrefix={arXiv},
      primaryClass={cs.LG},
      url={https://arxiv.org/abs/2110.14168}, 
}

@misc{hendrycks2021measuring,
      title={Measuring Massive Multitask Language Understanding}, 
      author={Dan Hendrycks and Collin Burns and Steven Basart and Andy Zou and Mantas Mazeika and Dawn Song and Jacob Steinhardt},
      year={2021},
      eprint={2009.03300},
      archivePrefix={arXiv},
      primaryClass={cs.CY},
      url={https://arxiv.org/abs/2009.03300}, 
}

@inproceedings{mihaylov2018can,
    title = "Can a Suit of Armor Conduct Electricity? A New Dataset for Open Book Question Answering",
    author = "Mihaylov, Todor  and
      Clark, Peter  and
      Khot, Tushar  and
      Sabharwal, Ashish",
    editor = "Riloff, Ellen  and
      Chiang, David  and
      Hockenmaier, Julia  and
      Tsujii, Jun{'}ichi",
    booktitle = "Proceedings of the 2018 Conference on Empirical Methods in Natural Language Processing",
    month = oct # "-" # nov,
    year = "2018",
    address = "Brussels, Belgium",
    publisher = "Association for Computational Linguistics",
    url = "https://aclanthology.org/D18-1260/",
    doi = "10.18653/v1/D18-1260",
    pages = "2381--2391"
}

@misc{kwon2023efficient,
      title={Efficient Memory Management for Large Language Model Serving with PagedAttention}, 
      author={Woosuk Kwon and Zhuohan Li and Siyuan Zhuang and Ying Sheng and Lianmin Zheng and Cody Hao Yu and Joseph E. Gonzalez and Hao Zhang and Ion Stoica},
      year={2023},
      eprint={2309.06180},
      archivePrefix={arXiv},
      primaryClass={cs.LG},
      url={https://arxiv.org/abs/2309.06180}, 
}

@misc{zoph2022stmoe,
      title={ST-MoE: Designing Stable and Transferable Sparse Expert Models}, 
      author={Barret Zoph and Irwan Bello and Sameer Kumar and Nan Du and Yanping Huang and Jeff Dean and Noam Shazeer and William Fedus},
      year={2022},
      eprint={2202.08906},
      archivePrefix={arXiv},
      primaryClass={cs.CL},
      url={https://arxiv.org/abs/2202.08906}, 
}

@misc{li2024mcsmoe,
      title={Merge, Then Compress: Demystify Efficient SMoE with Hints from Its Routing Policy}, 
      author={Pingzhi Li and Zhenyu Zhang and Prateek Yadav and Yi-Lin Sung and Yu Cheng and Mohit Bansal and Tianlong Chen},
      year={2024},
      eprint={2310.01334},
      archivePrefix={arXiv},
      primaryClass={cs.LG},
      url={https://arxiv.org/abs/2310.01334}, 
}

@misc{omi2025simbal,
      title={Load Balancing Mixture of Experts with Similarity Preserving Routers}, 
      author={Nabil Omi and Siddhartha Sen and Ali Farhadi},
      year={2025},
      eprint={2506.14038},
      archivePrefix={arXiv},
      primaryClass={cs.LG},
      url={https://arxiv.org/abs/2506.14038}, 
}

@misc{deng2025drpruning,
      title={DRPruning: Efficient Large Language Model Pruning through Distributionally Robust Optimization}, 
      author={Hexuan Deng and Wenxiang Jiao and Xuebo Liu and Jing Li and Min Zhang and Zhaopeng Tu},
      year={2025},
      eprint={2411.14055},
      archivePrefix={arXiv},
      primaryClass={cs.CL},
      url={https://arxiv.org/abs/2411.14055}, 
}

@inproceedings{
huang2025shapleymoe,
title={Discovering Important Experts for Mixture-of-Experts Models Pruning Through a Theoretical Perspective},
author={Weizhong Huang and Yuxin Zhang and Xiawu Zheng and Fei Chao and Rongrong Ji and Liujuan Cao},
booktitle={The Thirty-ninth Annual Conference on Neural Information Processing Systems},
year={2026},
url={https://openreview.net/forum?id=7kQjbCQwtT}
}

@misc{sakaguchi2021winogrande,
      title={WinoGrande: An Adversarial Winograd Schema Challenge at Scale}, 
      author={Keisuke Sakaguchi and Ronan Le Bras and Chandra Bhagavatula and Yejin Choi},
      year={2019},
      eprint={1907.10641},
      archivePrefix={arXiv},
      primaryClass={cs.CL},
      url={https://arxiv.org/abs/1907.10641}, 
}

@misc{rein2023gpqagraduatelevelgoogleproofqa,
  title={{GPQA}: A Graduate-Level Google-Proof Q\&A Benchmark},
  author={David Rein and Betty Li Hou and Asa Cooper Stickland and Jackson Petty and Richard Yuanzhe Pang and Julien Dirani and Julian Michael and Samuel R. Bowman},
  year={2023},
  eprint={2311.12022},
  archivePrefix={arXiv},
  primaryClass={cs.AI},
  url={https://arxiv.org/abs/2311.12022}
}

@misc{wang2024mmluprorobustchallengingmultitask,
  title={{MMLU-Pro}: A More Robust and Challenging Multi-Task Language Understanding Benchmark},
  author={Yubo Wang and Xueguang Ma and Ge Zhang and Yuansheng Ni and Abhranil Chandra and Shiguang Guo and Weiming Ren and Aaran Arulraj and Xuan He and Ziyan Jiang and Tianle Li and Max Ku and Kai Wang and Alex Zhuang and Rongqi Fan and Xiang Yue and Wenhu Chen},
  year={2024},
  eprint={2406.01574},
  archivePrefix={arXiv},
  primaryClass={cs.CL},
  url={https://arxiv.org/abs/2406.01574}
}

@misc{zhou2023instructionfollowingevaluationlargelanguage,
  title={Instruction-Following Evaluation for Large Language Models},
  author={Jeffrey Zhou and Tianjian Lu and Swaroop Mishra and Siddhartha Brahma and Sujoy Basu and Yi Luan and Denny Zhou and Le Hou},
  year={2023},
  eprint={2311.07911},
  archivePrefix={arXiv},
  primaryClass={cs.CL},
  url={https://arxiv.org/abs/2311.07911}
}
}

\appendix

\section{PPL Confound: Top-$k$-Then-Softmax Mechanism}
\label{app:ppl-confound}

gpt-oss uses \emph{topk-then-softmax} routing: the router selects the top-$k{=}4$ experts by raw logit, then applies softmax over only those $k$ values. With $\lambda_{\text{aux}}{=}0.9$, all 32 expert logits are nearly equal, making top-4 selection semi-random. After removing 8 experts ($32 \to 24$ at $r{=}0.25$), the smaller selection pool biases top-4 toward the truly strongest experts, improving per-token predictions. However, this PPL improvement comes at the cost of task-specific capabilities that PPL does not measure. Inter-method PPL comparisons (all at 24 experts) remain valid; we therefore focus on task accuracy for baseline comparisons throughout.

\section{Permutation Test Details}
\label{app:permtest}

The permutation test in Section~\ref{sec:frontier} evaluates whether the negative correlation between reasoning and knowledge task rankings is statistically valid or could arise by chance from arbitrary task groupings.

\textbf{Procedure.}\quad Let $\mathcal{T} = \{t_1, \ldots, t_{12}\}$ denote our 12 benchmark tasks, and let $a_{m,t}$ denote the accuracy of method $m$ on task $t$. For each possible partition of $\mathcal{T}$ into two equal-sized groups $(G_A, G_B)$ with $|G_A| = |G_B| = 6$, we compute the mean accuracy per method in each group: $\bar{a}_{m,A} = \frac{1}{6}\sum_{t \in G_A} a_{m,t}$ and similarly for $\bar{a}_{m,B}$. We then compute the Spearman rank correlation $\rho(G_A, G_B)$ between $\{\bar{a}_{m,A}\}_m$ and $\{\bar{a}_{m,B}\}_m$ across methods, and average over all $\binom{12}{6} = 924$ partitions to obtain the null distribution.

\textbf{Results (pruned methods only, $n=5$).}\quad The observed task clusters \{GSM8K, AIME25\} vs.\ \{MMLU, ARC-Challenge, ARC-Easy\} yield strongly negative average $\rho$. Of all 924 possible 3-vs-3 groupings from within these 6 core tasks (or equivalently, 6-vs-6 from the full 12 tasks), only 0.04\% produce correlations this negative, giving $p = 0.0004$.

\textbf{Results (including baseline, $n=6$).}\quad When the unpruned baseline is included as a sixth ``method,'' the average cross-group $\rho$ rises because the baseline dominates both cluster means, compressing the anti-correlation. The $p$-value remains significant, confirming that the ranking inversion among pruned methods is not masked by the baseline's overall superiority.

\textbf{Interpretation.}\quad The permutation test shows that the specific task clusters identified by PCA and $k$-means produce opposing method rankings; this is not a result of the small number of methods ($n=5$) or post-hoc task selection. The reasoning tasks (GSM8K, AIME25) and knowledge tasks (MMLU, ARC) shows fundamentally different expert importance profiles in over-dispersed models, validating the two-axis capability trade-off described in the main text.

\section{Extended Capability Frontier Analysis}
\label{app:frontier}

\textbf{Task clustering.}\quad $k$-means ($k=2$) on the task loading vectors produces a clean separation: \emph{Cluster~A} (knowledge/pattern-matching) contains MMLU, MMLU-Pro, ARC-Challenge, ARC-Easy, and PIQA; \emph{Cluster~B} (reasoning/generative) contains GSM8K (both), AIME25, GPQA, IFEval, WinoGrande, and OpenBookQA. These clusters are not imposed \emph{a priori} but emerge from the per-method accuracy patterns.

\textbf{Method rankings invert across clusters.}\quad Among pruned methods, the ranking on Cluster~B reasoning tasks is REAP $>$ MESA $>$ EvoESAP $>$ \textsc{RM-Uniform}, while on Cluster~A knowledge tasks it partially reverses: EvoESAP $>$ \textsc{RM-Uniform} $>$ MESA $>$ REAP. The Pearson correlation between the two cluster means across the pruned methods is strongly negative, indicating a near-perfect tradeoff. We caution that $n$ is small; the anti-correlation weakens when the baseline is included, because the unpruned model dominates both axes.

\textbf{GPQA-Diamond analysis.}\quad GPQA-Diamond requires \emph{both} graduate-level factual knowledge \emph{and} multi-step reasoning, placing it at the intersection of the two axes. It exhibits the second-largest accuracy range across pruning methods (21\%, after AIME25's 30\%): MESA achieves 65.48\% ($+1.0$\% over baseline), while REAP collapses to 47.72\% ($-16.8$\%).

\textbf{Mechanistic account.}\quad REAP scores experts by $\pi_i(x) \cdot \|\mathbf{E}_i(x)\|_2$. In over-dispersed models where $\pi_i \approx 1/E$, REAP scores are dominated by activation magnitude, selecting ``computation-heavy'' experts that contribute large output norms. These experts are disproportionately important for iterative computation (math reasoning) but not for distributional knowledge (factual recall). Conversely, router mass selects ``highway'' experts that handle high token volume. The key enabler of this divergence is the over-dispersed routing itself: because $\lambda_{\text{aux}} = 0.9$ decorrelates activation norms from routing frequencies, different scoring signals select different expert subsets. MESA achieves balance by optimizing worst-case routing perturbation across domains, avoiding the extreme of either axis.

\section{MESA Hyperparameter Ablations}
\label{app:ablations}

This section provides full ablation details for the calibration and hyperparameter results summarized in Section~\ref{sec:calibration}.

\subsection{Calibration Data Requirements}

\begin{table}[t]
\centering
\caption{Calibration sensitivity of MESA expert selection. Jaccard similarity
measures agreement with the reference configuration ($s{=}20$, $\ell{=}2048$)
on which experts to prune ($24 \times 8 = 192$ binary decisions at $r{=}0.25$).
Different calibration budgets explore different regions of the capability
Pareto frontier: lower calibration favors knowledge preservation while higher
calibration emphasizes domain fairness.}
\label{tab:calibration}
\small
\begin{tabular}{@{}l cc c@{}}
\toprule
\multicolumn{4}{c}{\textbf{(a) Samples per domain} (fixed $\ell=2048$)} \\
\midrule
Samples/domain & Jaccard $J$ & $\Delta$ experts & Character \\
\midrule
5   & 0.655 & 40 & Knowledge-biased \\
10  & 0.829 & 18 & Moderate \\
\rowcolor{gray!10}
20  & 1.000 & 0  & \checkmark Reference \\
50  & 0.979 & 2  & \checkmark Near-reference \\
100 & 0.901 & 10 & Domain-fair-biased \\
200 & 0.901 & 10 & Domain-fair-biased \\
\midrule
\multicolumn{4}{c}{\textbf{(b) Sequence length} (fixed $s=20$)} \\
\midrule
Seq.\ length & Jaccard $J$ & $\Delta$ experts & Character \\
\midrule
128  & 0.548 & 56 & High variance \\
256  & 0.627 & 44 & High variance \\
512  & 0.627 & 44 & High variance \\
\rowcolor{gray!10}
1024 & 1.000 & 0  & \checkmark Converged \\
2048 & 1.000 & 0  & \checkmark Reference \\
\bottomrule
\end{tabular}
\end{table}

Table~\ref{tab:calibration} shows MESA's sensitivity to calibration volume. We measure prune map agreement (Jaccard similarity) at $r{=}0.25$, where 192 binary expert decisions are made ($24$ layers $\times$ $8$ experts pruned per layer). The key finding is that \textbf{sequence length dominates sample count}: $\ell{\ge}1024$ tokens is necessary for stable routing statistics regardless of sample count, while $s{\ge}20$ samples suffices when sequence length is adequate. This aligns with the intuition that over-dispersed routing patterns require sufficient context to manifest stable domain-expert affinities.

When measured against a high-quality reference ($s{=}2000$, $l{=}8192$), the convergence dynamics are:
\begin{itemize}
\item $s{=}20$, $l{=}2048$: 64.1\% agreement (42 experts differ)
\item $s{=}200$, $l{=}2048$: 69.9\% agreement (34 experts differ)
\item $s{=}1000$, $l{=}4096$: 82.9\% agreement (18 experts differ)
\item $s{=}1000$, $l{=}8192$: 97.9\% agreement (only 2 experts differ)
\item $s{=}2000$, $l{=}4096$: 81.1\% agreement (20 experts differ)
\end{itemize}
The jump from $l{=}4096$ to $l{=}8192$ at $s{=}1000$ (+15.0\%) far exceeds the gain from $s{=}20$ to $s{=}200$ at $l{=}2048$ (+5.8\%), confirming that calibration compute should be invested in longer sequences rather than more samples.

Different calibration budgets produce divergent prune maps that explore different regions of the capability Pareto frontier rather than representing ``wrong'' selections. Higher calibration ($s{=}2000$, $l{=}8192$) shifts the trade-off toward domain balance (commonsense gains, PIQA +2\%) at the cost of knowledge-intensive tasks (MMLU-Pro $-$4\%). Lower calibration acts as implicit regularization, preventing over-optimization of the minimax objective and preserving broader knowledge. Practitioners should select calibration based on downstream priority: low calibration for knowledge preservation, high calibration for domain fairness.

\subsection{Iterations and Boost Rate}

\begin{table}[t]
\centering
\caption{MESA hyperparameter sensitivity at $r{=}0.25$. (a)~Increasing
iterations $T$ monotonically improves worst-domain loss and selection
agreement (Jaccard vs.\ $T{=}80$ reference). Convergence is effectively
reached by $T{=}15$. (b)~The boost rate $\eta_0$ controls convergence speed
vs.\ stability; values in $[0.05, 0.25]$ achieve near-identical solutions
with low oscillation.}
\label{tab:T-eta-ablation}
\small
\begin{tabular}{@{}l ccc@{}}
\toprule
\multicolumn{4}{c}{\textbf{(a) Iterations $T$} (fixed $\eta_0{=}0.15$)} \\
\midrule
$T$ & Worst-domain loss & Jaccard vs.\ $T{=}80$ & Best step \\
\midrule
1  & 0.0524 & 76.1\% & 0 \\
3  & 0.0509 & 81.1\% & 2 \\
5  & 0.0503 & 86.4\% & 4 \\
10 & 0.0492 & 92.0\% & 8 \\
\rowcolor{gray!10}
15 & 0.0491 & 95.9\% & 13 \\
20 & 0.0491 & 95.9\% & 13 \\
30 & 0.0491 & 97.9\% & 21 \\
40 & 0.0491 & 97.9\% & 21 \\
80 & 0.0491 & 100.0\% & 53 \\
\midrule
\multicolumn{4}{c}{\textbf{(b) Boost rate $\eta_0$} (fixed $T{=}20$)} \\
\midrule
$\eta_0$ & Worst-domain loss & Jaccard vs.\ default & Oscillation \\
\midrule
0.01 & 0.0502 & 79.4\% & 0.0004 \\
0.03 & 0.0494 & 90.1\% & 0.0005 \\
\rowcolor{gray!10}
0.05 & 0.0491 & 95.9\% & 0.0003 \\
0.08 & 0.0491 & 95.9\% & 0.0003 \\
0.10 & 0.0491 & 97.9\% & 0.0005 \\
0.15 & 0.0491 & 97.9\% & 0.0008 \\
0.20 & 0.0491 & 97.9\% & 0.0008 \\
0.25 & 0.0491 & 97.9\% & 0.0012 \\
0.30 & 0.0491 & 95.9\% & 0.0011 \\
0.40 & 0.0491 & 95.9\% & 0.0020 \\
0.60 & 0.0491 & 95.9\% & 0.0041 \\
\bottomrule
\end{tabular}
\end{table}

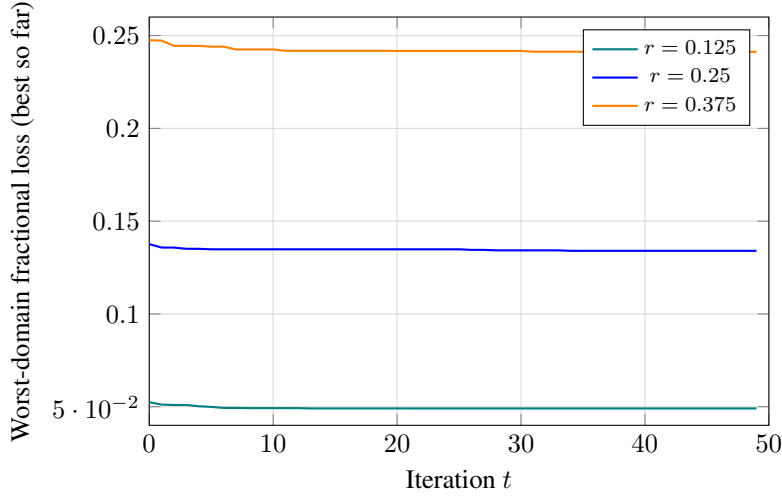
\begin{figure}[t]
\centering
%

\begin{tikzpicture}
\begin{axis}[
    width=0.7\textwidth,
    height=0.5\textwidth,
    xlabel={Iteration $t$},
    ylabel={Worst-domain fractional loss (best so far)},
    xmin=0, xmax=50,
    ymin=0.04, ymax=0.26,
    legend pos=north east,
    legend style={font=\footnotesize},
    grid=major,
    grid style={gray!30},
    every axis plot/.append style={thick},
]

\addplot[color=teal, mark=none] coordinates {
    (0, 0.052447) (1, 0.051169) (2, 0.050941) (3, 0.050941)
    (4, 0.050260) (5, 0.049928) (6, 0.049360) (7, 0.049360)
    (8, 0.049247) (9, 0.049247) (10, 0.049247) (11, 0.049247)
    (12, 0.049247) (13, 0.049087) (14, 0.049087) (15, 0.049087)
    (16, 0.049087) (17, 0.049087) (18, 0.049087) (19, 0.049087)
    (20, 0.049087) (21, 0.049082) (22, 0.049082) (23, 0.049082)
    (24, 0.049082) (25, 0.049082) (26, 0.049082) (27, 0.049082)
    (28, 0.049082) (29, 0.049082) (30, 0.049082) (31, 0.049082)
    (32, 0.049082) (33, 0.049082) (34, 0.049082) (35, 0.049082)
    (36, 0.049082) (37, 0.049082) (38, 0.049082) (39, 0.049082)
    (40, 0.049082) (41, 0.049082) (42, 0.049082) (43, 0.049082)
    (44, 0.049082) (45, 0.049082) (46, 0.049082) (47, 0.049082)
    (48, 0.049082) (49, 0.049082)
};
\addlegendentry{$r=0.125$}

\addplot[color=blue, mark=none] coordinates {
    (0, 0.137631) (1, 0.135772) (2, 0.135772) (3, 0.135102)
    (4, 0.135102) (5, 0.134844) (6, 0.134844) (7, 0.134844)
    (8, 0.134844) (9, 0.134844) (10, 0.134844) (11, 0.134844)
    (12, 0.134844) (13, 0.134844) (14, 0.134844) (15, 0.134844)
    (16, 0.134844) (17, 0.134844) (18, 0.134844) (19, 0.134844)
    (20, 0.134844) (21, 0.134844) (22, 0.134844) (23, 0.134844)
    (24, 0.134844) (25, 0.134844) (26, 0.134526) (27, 0.134526)
    (28, 0.134270) (29, 0.134270) (30, 0.134270) (31, 0.134270)
    (32, 0.134270) (33, 0.134270) (34, 0.134016) (35, 0.134016)
    (36, 0.134016) (37, 0.134016) (38, 0.134016) (39, 0.134016)
    (40, 0.134016) (41, 0.134016) (42, 0.134016) (43, 0.134016)
    (44, 0.134016) (45, 0.134016) (46, 0.134016) (47, 0.134016)
    (48, 0.134016) (49, 0.134016)
};
\addlegendentry{$r=0.25$}

\addplot[color=orange, mark=none] coordinates {
    (0, 0.247515) (1, 0.247281) (2, 0.244452) (3, 0.244452)
    (4, 0.244452) (5, 0.244054) (6, 0.244054) (7, 0.242528)
    (8, 0.242528) (9, 0.242528) (10, 0.242528) (11, 0.241804)
    (12, 0.241804) (13, 0.241804) (14, 0.241804) (15, 0.241804)
    (16, 0.241804) (17, 0.241804) (18, 0.241804) (19, 0.241804)
    (20, 0.241735) (21, 0.241735) (22, 0.241735) (23, 0.241735)
    (24, 0.241735) (25, 0.241735) (26, 0.241735) (27, 0.241735)
    (28, 0.241735) (29, 0.241735) (30, 0.241735) (31, 0.241304)
    (32, 0.241304) (33, 0.241304) (34, 0.241304) (35, 0.241304)
    (36, 0.241304) (37, 0.241304) (38, 0.241304) (39, 0.241304)
    (40, 0.241304) (41, 0.241304) (42, 0.241304) (43, 0.241304)
    (44, 0.241304) (45, 0.241304) (46, 0.241304) (47, 0.241304)
    (48, 0.241304) (49, 0.241304)
};
\addlegendentry{$r=0.375$}

\end{axis}
\end{tikzpicture}
\caption{MESA convergence for different pruning ratios $r$. The algorithm
converges within 6--20 iterations for $r \leq 0.375$. Higher ratios produce
higher loss plateaus but similar convergence dynamics, suggesting the
difficulty is in the objective landscape, not the optimizer.}
\label{fig:convergence}
\end{figure}

\textbf{Convergence dynamics.}\quad Figure~\ref{fig:convergence} shows the worst-domain fractional loss (best-so-far) as a function of iteration $t$ for multiple pruning ratios. At $r{=}0.125$ and $r{=}0.25$, convergence is rapid (within 6--12 steps). At $r{=}0.375$, convergence takes slightly longer (20 steps) but achieves a stable minimum. Higher pruning ratios produce higher loss plateaus, reflecting the fundamental difficulty of domain-fair pruning at aggressive ratios.

\textbf{Boost rate robustness.}\quad Table~\ref{tab:T-eta-ablation}(b) demonstrates broad robustness: any $\eta_0 \in [0.05, 0.60]$ achieves the same worst-domain loss (0.0491) at $r{=}0.25$, with Jaccard agreement $\geq$95.9\% against the default. Oscillation amplitude increases with $\eta_0$ (from 0.0003 to 0.0041) but the best-snapshot mechanism ensures this does not degrade the final solution. We recommend $\eta_0 = 0.15$ as a safe default.

\textbf{Decay schedules.}\quad Linear, cosine, and constant decay all achieve identical final losses (within numerical precision), converging in 8--17 steps. The default linear schedule ($\eta_t = \eta_0 \cdot (1 - 0.5 \cdot t/(T-1))$) decays to 50\% of the initial rate, providing coarse-to-fine adjustment.

\section{Routing Structure Visualization}
\label{app:routing-viz}

\subsection{Per-Domain Routing Mass}

\begin{figure}[h]
\centering
\includegraphics[width=\textwidth]{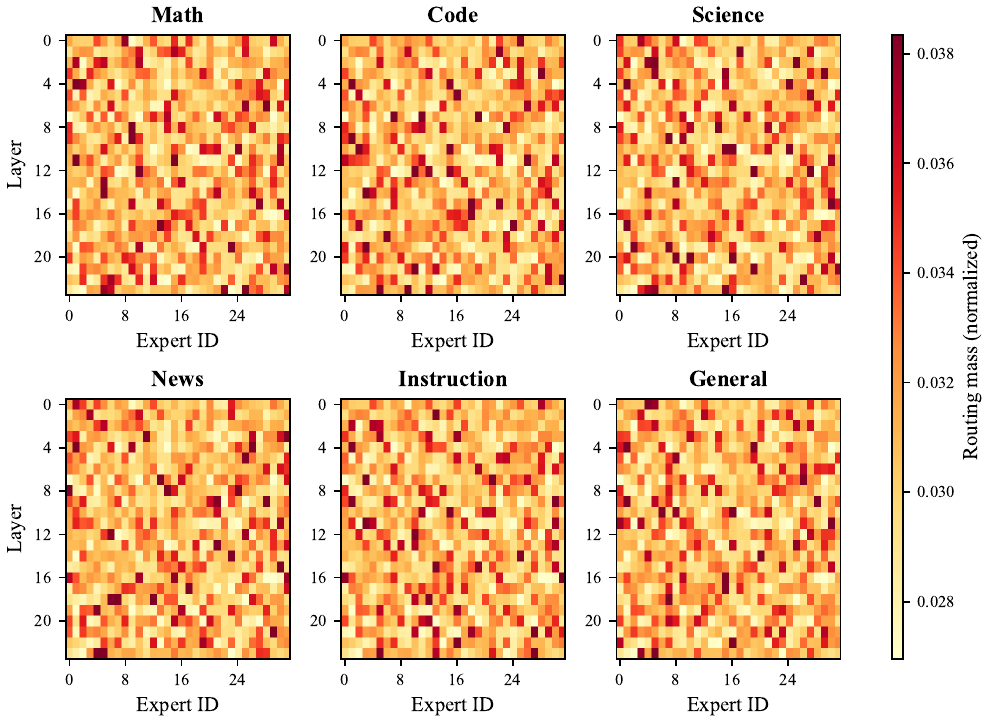}
\caption{Per-domain routing mass across all 24 layers and 32 experts in
gpt-oss-20b. Each cell shows the fraction of routing probability mass that
domain $d$ assigns to expert $e$ at layer $l$. The narrow colorbar range
(0.028--0.038 vs.\ the uniform expectation of $1/32 \approx 0.031$) confirms
over-dispersion: no single expert dominates for any domain. However, subtle
per-domain structure is visible; certain experts receive slightly elevated
mass for specific domains (e.g., experts 8--12 in layers 16--20 for Code).
These weak but consistent signals are exactly what MESA's domain-conditional
scoring exploits: individually negligible, they become collectively meaningful
when the minimax objective accumulates evidence across all calibration tokens.}
\label{fig:routing-heatmap}
\end{figure}

The heatmaps in Figure~\ref{fig:routing-heatmap} illustrate why global importance metrics fail in the over-dispersed regime. In a normally-routed model like Mixtral, equivalent heatmaps would show sharp contrast (clear hot-spots at ${\sim}0.2$--$0.4$ routing mass with most cells near zero). Here, the entire grid appears near-uniform, with domain-specific variation compressed into a 0.01 dynamic range. A global importance score averages across domains and layers, cancels out these weak signals entirely. MESA's per-domain measurement preserves them.

\subsection{Cross-Domain Expert Overlap}

\begin{figure}[h]
\centering
\includegraphics[width=0.65\textwidth]{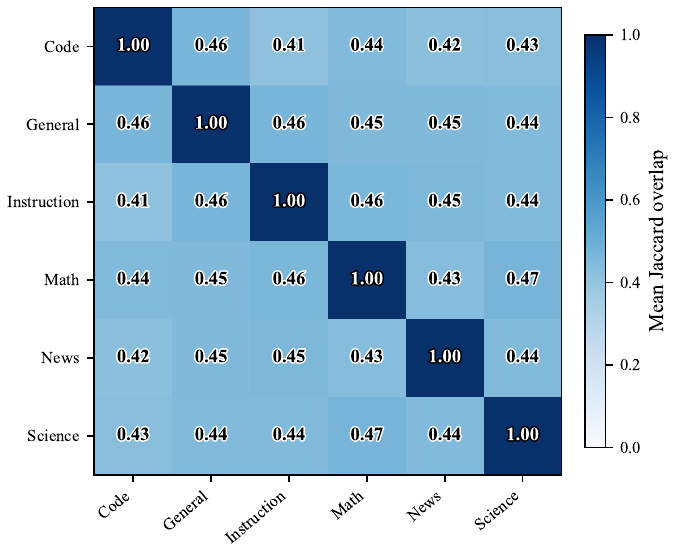}
\caption{Jaccard overlap between domains' top-8 most-routed experts per layer,
averaged across all 24 layers. High off-diagonal values (0.41--0.47) confirm
that domains share nearly half their preferred experts due to over-dispersed
routing. The slight asymmetries (e.g., Math--Science 0.47 vs.\ Code--Instruction
0.41) reflect residual domain structure that domain-fair scoring can exploit.}
\label{fig:overlap-matrix}
\end{figure}

Figure~\ref{fig:overlap-matrix} quantifies the challenge for domain-fair pruning. When the top-8 experts for any two domains overlap by ${\sim}45\%$, pruning an expert that appears important for one domain affects another. This high baseline overlap explains why naive per-domain scoring (keeping each domain's top experts separately) is insufficient: it would produce nearly identical selections for all domains. MESA's iterative minimax approach navigates this overlap by identifying the marginal experts whose removal disproportionately harms one domain relative to others, precisely those in the 0.41--0.47 overlap zone where domains partially diverge.

\subsection{Per-Task Degradation Patterns}

\begin{figure*}[h]
\centering
\includegraphics[width=\textwidth]{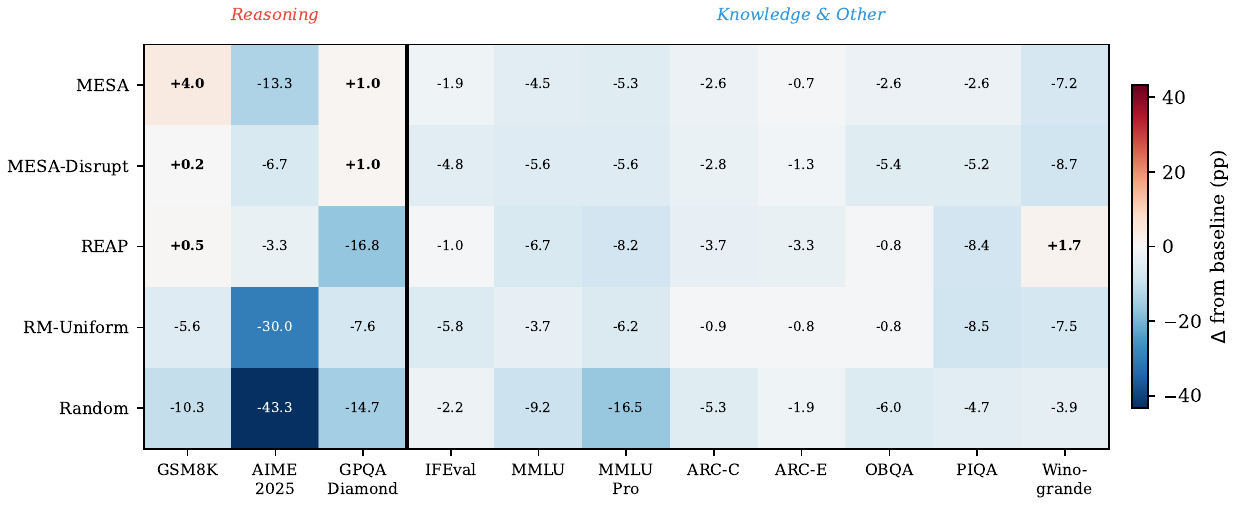}
\caption{Per-task accuracy change from baseline at $r=0.25$ for all evaluated
pruning methods. The heatmap reveals structured degradation patterns: reasoning
tasks (left columns) show large variance across methods (AIME25 spans $-3.3$ to
$-43.3$\%), while knowledge tasks (right columns) show uniformly mild
degradation ($<5$\% for most methods). MESA achieves the lightest overall
shading (smallest absolute degradation across all tasks), while REAP shows
a distinctive pattern of severe GPQA loss ($-16.8$\%) alongside strong
reasoning preservation.}
\label{fig:delta-heatmap}
\end{figure*}

Figure~\ref{fig:delta-heatmap} shows per-task accuracy changes from baseline for all methods at $r=0.25$. The structured degradation pattern confirms that the capability trade-off is task-specific rather than uniform across benchmarks.

\section{Regime Transition: Importance Signal Degradation}
\label{app:regime-transition}

\begin{figure}[t]
\centering
\includegraphics[width=0.95\columnwidth]{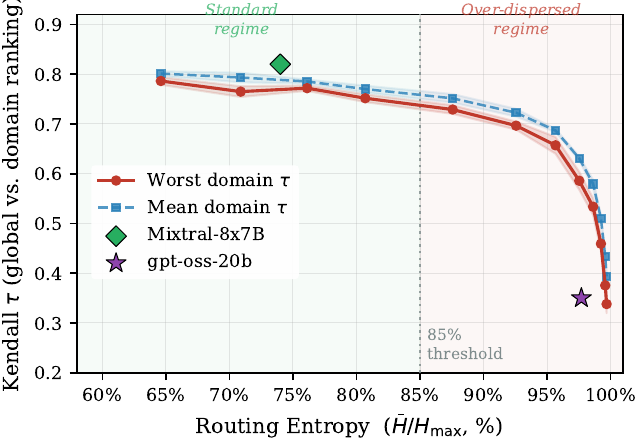}
\caption{Routing entropy determines importance signal quality. As routing
entropy increases toward uniformity, Kendall-$\tau$ between global importance
ranking and worst-domain ranking decreases sharply. The 85\% band
(dashed) is a practical diagnostic, not a sharp physical boundary, and
separates reliable (Mixtral, 74.3\%) from degraded (gpt-oss,
90.8\%) regimes. Simulated on 24-layer, 32-expert matrices with 6
calibration domains; real data points marked with stars.}
\label{fig:entropy-transition}
\end{figure}

We simulate routing matrices at varying entropy levels to characterize the transition between standard and over-dispersed regimes. Figure~\ref{fig:entropy-transition} shows how global importance signal quality degrades as routing entropy increases.

\section{Qualitative Analysis: Knowledge Degradation Under Pruning}
\label{app:capability-cliff}

\begin{figure*}[t]
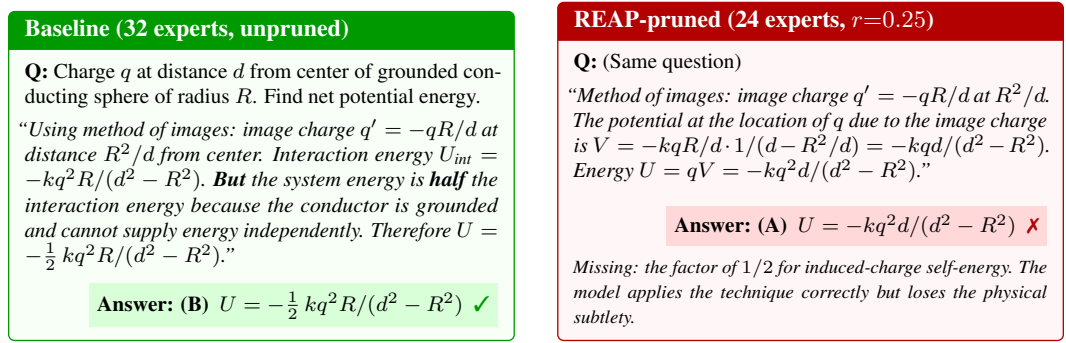

\centering
\begin{minipage}[t]{0.48\textwidth}
\begin{tcolorbox}[
  colback=green!3!white,
  colframe=green!60!black,
  title={\textbf{Baseline} (32 experts, unpruned)},
  fonttitle=\small\bfseries,
  coltitle=white,
  colbacktitle=green!60!black,
  boxrule=0.5pt,
  arc=2pt,
  left=3pt, right=3pt, top=2pt, bottom=2pt
]
\footnotesize
\textbf{Q:} Charge $q$ at distance $d$ from center of grounded conducting sphere of radius $R$. Find net potential energy.

\vspace{3pt}
\textit{``Using method of images: image charge $q' = -qR/d$ at distance $R^2/d$ from center. Interaction energy $U_{\text{int}} = -kq^2R/(d^2-R^2)$. \textbf{But} the system energy is \textbf{half} the interaction energy because the conductor is grounded and cannot supply energy independently. Therefore $U = -\tfrac{1}{2}\,kq^2 R/(d^2-R^2)$.''}

\vspace{4pt}
\hfill\colorbox{green!15}{\textbf{Answer: (B)}\enspace$U = -\tfrac{1}{2}\,kq^2 R/(d^2-R^2)$\enspace\textcolor{green!60!black}{\cmark}}
\end{tcolorbox}
\end{minipage}%
\hfill%
\begin{minipage}[t]{0.48\textwidth}
\begin{tcolorbox}[
  colback=red!3!white,
  colframe=red!70!black,
  title={\textbf{REAP-pruned} (24 experts, $r{=}0.25$)},
  fonttitle=\small\bfseries,
  coltitle=white,
  colbacktitle=red!70!black,
  boxrule=0.5pt,
  arc=2pt,
  left=3pt, right=3pt, top=2pt, bottom=2pt
]
\footnotesize
\textbf{Q:} (Same question)

\vspace{3pt}
\textit{``Method of images: image charge $q' = -qR/d$ at $R^2/d$. The potential at the location of $q$ due to the image charge is $V = -kqR/d \cdot 1/(d - R^2/d) = -kqd/(d^2-R^2)$. Energy $U = qV = -kq^2 d/(d^2-R^2)$.''}

\vspace{8pt}
\hfill\colorbox{red!15}{\textbf{Answer: (A)}\enspace$U = -kq^2 d/(d^2-R^2)$\enspace\textcolor{red!70!black}{\xmark}}

\vspace{4pt}
{\scriptsize \textit{Missing: the factor of $1/2$ for induced-charge self-energy. The model applies the technique correctly but loses the physical subtlety.}}
\end{tcolorbox}
\end{minipage}

\caption{\textbf{Knowledge degradation on GPQA-Diamond.} Same model (gpt-oss-20b), same graduate-level electrostatics problem.
\emph{Left:} the unpruned model correctly identifies that system energy is half the interaction energy for a grounded conductor.
\emph{Right:} REAP-pruned model sets up the method of images identically but omits the factor of $1/2$, a specific physical concept lost during pruning. This pattern, correct technique but missing conceptual knowledge, explains REAP's 17.8\% GPQA gap relative to MESA: the pruned model retains procedural skills but loses domain-specific physical reasoning stored in the removed experts.}
\label{fig:capability-cliff}
\end{figure*}

Aggregate accuracy (e.g., REAP at 47.72\% on GPQA) masks the nature of individual failures. Figure~\ref{fig:capability-cliff} shows a graduate-level electrostatics problem where the baseline and REAP-pruned model both set up the method of images correctly, but REAP fails to apply a specific physical subtlety (the factor of 1/2 for induced-charge energy), selecting the wrong formula.

\section{Gemma-4 Temperature-0 Replication}
\label{app:gemma-temp0}

The Gemma-4-26B-A4B-it flagship result reported in the main text (Table~\ref{tab:results-gemma}) uses
temperature $0.6$, matching the decoding protocol used for every other result in this paper (which
preserves chain-of-thought). To confirm that the worst-domain protection result is not an artifact of
sampling temperature, Table~\ref{tab:gemma-temp0} below reports an independent replication of the same
$r{=}0.25$ head-to-head under temperature $0$ (deterministic decoding), with 3-seed mean $\pm$ std, on
the subset of tasks for which temperature-0 replication data is available. We disclose both
temperatures rather than reporting only the more favorable one.

\begin{table}[t]
\centering
\caption{Gemma-4-26B-A4B-it $r=0.25$ head-to-head at temperature $0$ (deterministic decoding), mean
$\pm$ std over 3 seeds, on the 5 tasks with temperature-$0$ replication data. Best pruned result per
task in \textbf{bold}; non-overlapping intervals on GPQA-Diamond. This is an independent replication
of the main-text temperature-$0.6$ result (Table~\ref{tab:results-gemma}) under deterministic
decoding, disclosed alongside it rather than in place of it. Additional benchmarks (MMLU, MMLU-Pro,
IFEval, PIQA, OpenBookQA, WinoGrande) are pending temperature-$0$ replication.}
\label{tab:gemma-temp0}
\footnotesize
\setlength{\tabcolsep}{5pt}
\begin{tabular}{@{}l cccc@{}}
\toprule
Task & Unpruned & MESA (ours) & REAP & \textsc{RM-Uni.} \\
\midrule
AIME'25 & 86.67{\tiny$\pm$3.33} & \textbf{90.00}{\tiny$\pm$3.33} & 76.67{\tiny$\pm$4.71} & 70.00{\tiny$\pm$5.44} \\
GSM8K & 96.51{\tiny$\pm$0.38} & \textbf{95.98}{\tiny$\pm$0.45} & 93.18{\tiny$\pm$0.62} & 89.54{\tiny$\pm$0.88} \\
GPQA-Diamond & 74.24{\tiny$\pm$1.21} & \textbf{65.66}{\tiny$\pm$1.82} & 55.56{\tiny$\pm$1.95} & 48.99{\tiny$\pm$2.15} \\
ARC-Challenge & 96.84{\tiny$\pm$0.42} & \textbf{95.56}{\tiny$\pm$0.51} & 92.40{\tiny$\pm$0.68} & 88.15{\tiny$\pm$0.82} \\
ARC-Easy & 98.82{\tiny$\pm$0.25} & \textbf{98.23}{\tiny$\pm$0.31} & 96.50{\tiny$\pm$0.45} & 93.80{\tiny$\pm$0.60} \\
\bottomrule
\end{tabular}
\vspace{2pt}

{\scriptsize A domain-aware REAP control (REAP given the same per-domain conditioning as
MESA) reaches GPQA-Diamond $58.4$ (vs.\ MESA's $65.66$ at temperature $0$ / $67.85$ at
temperature $0.6$), showing MESA's gain comes from the minimax worst-case \emph{objective}, not
merely from conditioning REAP's importance scores by domain.}
\end{table}

\section{gpt-oss-120B at $r=0.5$: Scale and Aggressive-Ratio Generality}
\label{app:gptoss120b}

As a complementary generality check within the same over-dispersed family, we scale from
gpt-oss-20B to \textbf{gpt-oss-120B} (roughly $5\times$ larger, same $\lambda_{\text{aux}}=0.9$
routing regime) and test a much more aggressive pruning ratio, $r=0.5$. Table~\ref{tab:results-120b}
reports the same 11-benchmark comparison across MESA, REAP, \textsc{RM-Uniform}, and the unpruned
model. \textsc{RM-Uniform} collapses entirely (all-task $0.00$), a genuine failure of that method at
this scale and ratio, not a missing measurement. Between MESA and REAP the result is mixed: MESA wins
the worst domain (GPQA-Diamond, $58.46$ vs.\ $51.85$) along with AIME, IFEval, MMLU, OpenBookQA, and
PIQA, while REAP wins on GSM8K, ARC-Challenge, ARC-Easy, MMLU-Pro, and WinoGrande. We present this
plainly as an aggressive-pruning stress test rather than a sweep: MESA's worst-domain protection holds
even at $r=0.5$, though it does not dominate every benchmark at this ratio. This is consistent with
the minimax reframing used throughout the paper. We make no claim of average-case superiority here,
only that the worst-case guarantee the method targets continues to hold under substantially heavier
pruning and at larger scale.

\begin{table*}[t]
\centering
\caption{Task accuracy (\%) on \textbf{gpt-oss-120B at $r=0.5$} (aggressive expert pruning,
$128 \to 64$ experts per layer), single measurement per cell.
Best pruned result per task in \textbf{bold}. \textsc{RM-Uniform} collapses to $0.00$ on every task at
this ratio, a genuine failure of the method at $r=0.5$ rather than a missing measurement. On the
worst domain (GPQA-Diamond, the hardest task in the suite), MESA leads REAP by $6.6$pp; MESA also
leads on AIME, IFEval, MMLU, OpenBookQA, and PIQA. REAP edges ahead on GSM8K, ARC-Challenge,
ARC-Easy, MMLU-Pro, and WinoGrande. We report this honestly as a \emph{mixed} outcome at this aggressive ratio: MESA's
worst-domain protection holds, but it does not dominate every benchmark under such heavy pruning.}
\label{tab:results-120b}
\footnotesize
\setlength{\tabcolsep}{3.5pt}
\resizebox{\textwidth}{!}{%
\begin{tabular}{@{}l ccc cccc cccc@{}}
\toprule
& \multicolumn{3}{c}{Math Reasoning} & \multicolumn{4}{c}{Knowledge \& Commonsense} & \multicolumn{4}{c}{Multi-domain \& Instruction} \\
\cmidrule(lr){2-4} \cmidrule(lr){5-8} \cmidrule(lr){9-12}
Method & AIME'25 & GSM8K & GPQA & ARC-C & ARC-E & OBQA & PIQA & MMLU & MMLU-Pro & WinoGr. & IFEval \\
\midrule
Unpruned & 74.00 & 95.63 & 70.36 & 96.38 & 98.67 & 93.91 & 84.93 & 87.76 & 77.78 & 81.03 & 86.50 \\
MESA (ours) & \textbf{72.22} & 95.58 & \textbf{58.46} & 92.38 & 95.88 & \textbf{93.56} & \textbf{77.95} & \textbf{83.01} & 64.20 & 77.27 & \textbf{87.15} \\
REAP & 48.89 & \textbf{96.26} & 51.85 & \textbf{95.14} & \textbf{98.18} & 93.33 & 77.93 & 82.05 & \textbf{68.96} & \textbf{81.16} & 84.50 \\
\textsc{RM-Uni.} & 0.00 & 0.00 & 0.00 & 0.00 & 0.00 & 0.00 & 0.00 & 0.00 & 0.00 & 0.00 & 0.00 \\
\bottomrule
\end{tabular}%
}
\vspace{2pt}
{\scriptsize \textsc{RM-Uniform} collapses completely at $r=0.5$ on gpt-oss-120B (all-task $0.00$); this
is a real degeneracy of the method under aggressive pruning at this scale, not an unmeasured cell.
Single-seed; a multi-seed confirmation is future work.}
\end{table*}

\end{document}